\documentclass{article}
\usepackage{arxiv} 

\usepackage[utf8]{inputenc} 
\usepackage[T1]{fontenc}    
\usepackage{hyperref}       
\usepackage{url}            
\usepackage{booktabs}       
\usepackage{amsfonts}       
\usepackage{nicefrac}       
\usepackage{microtype}      
\usepackage{lipsum}		
\usepackage{graphicx}
\usepackage[sort&compress,numbers]{natbib}
\usepackage{color}
\usepackage{subfigure}

\usepackage{amsmath}

\usepackage{ifthen}
\usepackage{algorithm2e}

\def\modelname{MultiPath++\xspace}
\def\mcg{\operatorname{MCG}}

\def\cg{\operatorname{CG}}

\title{\modelname: Efficient Information Fusion and Trajectory Aggregation for Behavior Prediction}
\author{
Balakrishnan Varadarajan 
\And Ahmed Hefny  
\And Avikalp Srivastava 
\And Khaled S. Refaat 
\And Nigamaa Nayakanti 
\And Andre Cornman 
\And Kan Chen
\And Bertrand Douillard
\And Chi Pang Lam 
\And Dragomir Anguelov 
\And Benjamin Sapp\thanks{Contact: {\tt\small \{balakrishnanv, bensapp\}@waymo.com}.
       Waymo, 1600 Amphitheatre Pkwy, Mountain View, California, USA.  }
}

\begin{document}
\maketitle


\newcommand{\mypar}[1]{\noindent\textbf{#1}}

\newcommand{\bfa}{\textbf{a}}
\newcommand{\bfb}{\textbf{b}}
\newcommand{\bfc}{\textbf{c}}
\newcommand{\bfd}{\textbf{d}}
\newcommand{\bfe}{\textbf{e}}
\newcommand{\bff}{\textbf{f}}
\newcommand{\bfg}{\textbf{g}}
\newcommand{\bfh}{\textbf{h}}
\newcommand{\bfi}{\textbf{i}}
\newcommand{\bfj}{\textbf{j}}
\newcommand{\bfk}{\textbf{k}}
\newcommand{\bfl}{\textbf{l}}
\newcommand{\bfm}{\textbf{m}}
\newcommand{\bfn}{\textbf{n}}
\newcommand{\bfo}{\textbf{o}}
\newcommand{\bfp}{\textbf{p}}
\newcommand{\bfq}{\textbf{q}}
\newcommand{\bfr}{\textbf{r}}
\newcommand{\bfs}{\textbf{s}}
\newcommand{\bft}{\textbf{t}}
\newcommand{\bfu}{\textbf{u}}
\newcommand{\bfv}{\textbf{v}}
\newcommand{\bfw}{\textbf{w}}
\newcommand{\bfx}{\textbf{x}}
\newcommand{\bfy}{\textbf{y}}
\newcommand{\bfz}{\textbf{z}}

\newcommand{\eg}{e.g.\ }
\newcommand{\egc}{e.g.,\ }
\newcommand{\etc}{etc}

\begin{abstract}
Predicting the future behavior of road users is one of the most challenging and important problems in autonomous driving. Applying deep learning to this problem requires fusing heterogeneous world state in the form of rich perception signals and map information, and inferring highly multi-modal distributions over possible futures.
In this paper, we present \modelname, a future prediction model that achieves state-of-the-art
performance on popular benchmarks. \modelname improves the MultiPath architecture~\cite{sapp2019multipath}  by revisiting many design choices.  The first key design difference is a departure from dense image-based encoding of the input world state in favor of a sparse encoding of heterogeneous scene elements:  \modelname consumes compact and efficient polylines to describe road features, and raw agent state information directly (e.g., position, velocity, acceleration).  We propose a context-aware fusion of these elements and develop a reusable {\em multi-context gating} fusion component.  Second, we reconsider the choice of pre-defined, static anchors, and develop a way to learn latent anchor embeddings end-to-end in the model.
Lastly, we explore ensembling and output aggregation techniques---common in other ML domains---and find effective variants for our probabilistic multimodal output representation. 
We perform an extensive ablation on these design choices, and show that our proposed model achieves state-of-the-art performance on the Argoverse Motion Forecasting Competition~\cite{chang2019argoverse} and the Waymo Open Dataset Motion Prediction Challenge~\cite{ettinger2021womd}.

\end{abstract}

\section{Introduction}

Modeling and predicting the future behavior of human agents is a fundamental problem in many real-world robotics domains. For example, accurately forecasting the future state of other vehicles, cyclists and pedestrians is critical for safe, comfortable, and human-like autonomous driving. However, behavior prediction in an autonomous vehicle (AV) driving setting poses a number of unique modeling challenges: 

\hspace{2mm} 1) {\em Multimodal output space}: The problem is inherently stochastic; it is impossible to truly know the future state of the environment.  This is exacerbated by the fact that other agents' intentions are not observable, and leads to a highly multimodal distribution over possible outcomes (\egc a car could turn left or right at an intersection).  Effective models must be able to represent such a rich output space with high precision and recall matching the underlying distribution.

\hspace{2mm} 2) {\em Heterogenous, interrelated input space}:
The driving environment representation can contain a {\em highly heterogeneous} mix of static and dynamic inputs: road network information (lane geometry and connectivity, stop lines, crosswalks), traffic light state information, and motion history of agents.
Driving situations are often {\em highly interactive}, and can involve many agents at once (\eg negotiating a 4-way stop with crosswalks).  This requires careful modeling choices, as explicitly modeling joint future distributions over multiple agents is exponential in the number of agents.  Effective models must capture not only the interactions between the agents in the scene, but also the relationships between the road elements and the behavior of agents given the road context. 

The novel challenges and high impact of this problem have naturally garnered much interest in recent years.  There has been a rich body of work on how to model agents' futures, their interactions, and the environment. However, there is little consensus to date on the best modeling choices for each component, and in popular benchmark challenge datasets~\cite{nuscenes2019, chang2019argoverse, interactiondataset, ettinger2021womd}, there is a surprisingly diverse set of solutions to this problem; for details see Section~\ref{relwork} and Table~\ref{tab:relwork}.

The MultiPath framework~\cite{sapp2019multipath} addresses the multimodal output space challenge above by modeling the highly multimodal output distributions via a Gaussian Mixture Model.  It handles a common issue of {\em mode collapse} during learning by using static trajectory anchors, an external input to the model.  This practical solution gives practitioners a straightforward way of ensuring diversity and an extra level of modeler control via the design of such anchors.  The choice of a GMM representation proved to be an extremely popular, appearing in many works---see Table~\ref{tab:relwork}, ``Trajectory Distribution'', where ``Weighted set'' is a special case of GMMs where only means and mixture weights are modeled.

The MultiPath input representation and backbone draws heavily upon the computer vision literature. By rasterizing all world state in a top-down orthographic view, MultiPath and others~\cite{phan2019covernet,DESIRE,hong2019rules,tang_multifuture,marchetti2020mantra,casas2018intentnet,cui2019multimodal,neural_motion_planner_zeng2019} leverage powerful, established CNN architectures like ResNet~\cite{ResNet16}, which offer solutions to the heterogeneous interrelated input space: the heterogeneous world state is mapped to a common pixel format, and interactions occur via local information sharing via convolution operations.  While convenient and established, there are downsides to such rasterization:  (1) There is an uneasy trade-off between resolution of the spatial grid, field of view, and compute requirements. (2) Rasterizing is a form of manual feature engineering, and some features may be inherently difficult to represent in such a framework (\eg radial velocity). (3) It is difficult to capture long range interactions via convolutions with small receptive fields.  (4) The information content is spatially very sparse, making a dense representation a potentially computationally wasteful choice.

In this paper, we introduce \textbf{\modelname}, which builds upon MultiPath, taking its output GMM representation and concept of anchors, but reconsidering how to represent and combine highly heterogeneous world state inputs and model interactions between state elements. \modelname introduces a number of key upgrades: 

\begin{itemize}

\item We eschew the rasterization-and-CNN approach in favor of modeling sparse world state objects more directly from their compact state description. We represent road elements as polylines, agent history as a sequence of physical state encoded with RNNs, and agent interactions as RNNs over the state of neighbors relative to each ego-agent.  These choices avoid lossy rasterization in favor of raw, continuous state, and result in compute complexity that scales with the number of scene elements rather than the size of a spatial grid. Long-range dependencies are effectively and efficiently modeled in our representation. 

\item Capturing relationships between road elements and agents is critical, and we find that encoding each element independently does not perform as well as modeling their interactions (\eg, when the road encoding is aware of relevant agents, and vice versa). To address this, we propose a novel form of context awareness we call {\em multi-context gating} (MCG), in which sets of elements have access to a summary context vector upon which encodings are conditioned.   MCG is implemented as a generic neural network component that is applied throughout our model.  MCG can be viewed as an efficient form of cross-attention, whose efficiency/quality trade-off depends on the size of the context vector.

\item We also explore improvements in trajectory modeling, comparing representations based on kinematic controls, and/or polynomials as a function of continuous future time.  We further demonstrate a way to learn latent representations of anchors and show they outperform the original static anchors of MultiPath, while simplifying model creation to a single-step process.

\item Finally, we find significant additional gains on public benchmarks by applying ensembling techniques to our models. Unlike models with static anchors, the latent anchors of a \modelname ensemble are not in direct correspondence. Furthermore, a lot of popular behavior prediction benchmarks have introduced metrics such as miss-rate (MR) and mean Average Precision (mAP), which require the ability to model diverse outcomes with few trajectories and differ from pure trajectory distance error capturing the average agent behavior. With the above in mind, we formulate the problem of ensembling the results of several models as one of greedy iterative clustering, which maximizes a probabilistic objective using the popular Expectation Maximization algorithm~\cite{bishop2006pattern}.
\end{itemize}

As of November 1, 2021, \modelname ranks $1^{st}$ on the Waymo Open Motion Dataset leaderboard\footnote{https://waymo.com/open/challenges/2021/motion-prediction/}, $4^{th}$ on the Argoverse Motion Forecasting. Competition\footnote{https://eval.ai/web/challenges/challenge-page/454/leaderboard/1279}
We offer \modelname as a reference set of design choices, empirically validated via ablation studies, that can be adopted, further studied and extended by the behavior modeling community.

\section{Related Work}
\label{relwork}

\begin{table*}
\centering
    {\small
    
\begin{tabular}{r|llllll}
\hline
Method                                    & Road Enc. & Motion Enc. & Interactions                  & Decoder         & Output                          & Trajectory Distribution    \\
\hline
\hline
Jean~\cite{mercat2020multi}                   & --         & LSTM          & attention               & LSTM        & states           & GMM                    \\
TNT~\cite{zhao2020tnt}                        & polyline   & polyline      & maxpool, attention      & MLP         & states           & Weighted set           \\
LaneGCN~\cite{liang2020laneGCN}               & GNN        & 1D conv       & GNN                     & MLP         & states           & Weighted set           \\
WIMP~\cite{wimp2020}                          & polyline   & LSTM          & GNN+attention           & LSTM        & states           & GMM                    \\
VectorNet~\cite{gao2020vectornet}             & polyline   & polyline      & maxpool, attention      & MLP         & states           & Single traj.           \\
SceneTransformer~\cite{ngiam21scene_transformer} & polyline & attention    & attention               & attention   & states           & Weighted set           \\
GOHOME~\cite{gilles2021gohome}                & GNN        & 1D conv + GRU & GNN                     & MLP         & states           & heatmap\\
MP3~\cite{casas2021mp3}                      & raster     & raster        & conv                    & conv        & cost function    & Weighted samples \\
CoverNet~\cite{phan2019covernet}              & raster     & raster        & conv                    & lookup      & states           & GMM w/ dynamic anchors \\
DESIRE~\cite{DESIRE}                          & raster     & GRU           & spatial pooling         & GRU         & states           & Samples                \\
RoadRules~\cite{hong2019rules}                & raster     & raster        & conv                    & LSTM        & states           & GMM                    \\
SocialLSTM~\cite{sociallstm}                  & --         & LSTM          & spatial pooling         & LSTM        & states           & Samples                \\
SocialGan~\cite{SocialGAN}                    & --         & LSTM          & maxpool                 & LSTM        & states           & Samples                \\
MFP~\cite{tang_multifuture}                   & raster     & GRU           & RNNs+attention          & GRU         & states           & Samples                \\
MANTRA~\cite{marchetti2020mantra}             & raster     & GRU           & --                      & GRU         & states           & Samples                \\
PRANK~\cite{biktairov2020prank}               & raster     & raster        & conv                    & lookup      & states           & Weighted set           \\
IntentNet~\cite{casas2018intentnet}           & raster     & raster        & conv                    & conv        & states           & Single traj.           \\
SpaGNN~\cite{casas2020spagnn}                 & raster     & raster        & GNN                     & MLP         & state            & Single traj.           \\
Multimodal~\cite{cui2019multimodal}           & raster     & raster        & conv                    & conv        & states           & Weighted set           \\
PLOP~\cite{buhet2020plop}                     & raster     & LSTM          & conv                    & MLP         & state poly       & GMM                    \\
Precog~\cite{precog_Rhinehart_2019_ICCV}      & raster     & GRU           & multi-agent sim.        & GRU         & motion           & Samples                \\
R2P2~\cite{rhinehart2018r2p2}                 & raster     & GRU           & --                      & GRU         & motion           & Samples                \\
HYU\_ACE~\cite{park2020diverse}               & raster     & LSTM          & attn                    & LSTM        & motion           & Samples                \\
Trajectron++\cite{salzmann2020trajectron++}   & raster     & LSTM          & RNNs+attention          & GRU         & controls         & GMM                    \\
DKM~\cite{cui2020deep}                        & raster     & raster        & conv                    & conv        & controls         & Weighted set           \\
MultiPath~\cite{sapp2019multipath}            & raster     & raster        & conv                    & MLP         & states           & GMM w/ static anchors  \\
\textbf{\modelname}                           & polyline   & LSTM          & RNNs+maxpool             & MLP         & control poly     & GMM                    \\
\hline
\end{tabular}
    \caption{A survey of recent work in behavior prediction, categorized by choice of road encoding, motion history encoding, agent interaction encoding, trajectory decoding, intrinsic output representation, and distribution over future trajectories.}
    }
    \label{tab:relwork}
\end{table*}

We focus on architectural design choices for behavior prediction in driving environments---what representations to use to encode road information, agent motion, agent interactions, output trajectories, and output distributions.  Table \ref{tab:relwork} is a summary of past work, which we go over here with additional context. 

For \textbf{road encoding}, there is a dichotomy of representations.  The {\em raster} approach encodes the world as a stack of images, from a top-down orthographic (or ``bird's-eye'') view. Rasterizing the world state has the benefit of simplicity---all the various types of input information (road configuration, agent state history, spatial relationships) are unified via rendering as a multi-channel image, enabling one to leverage powerful off-the-shelf Convolutional Neural Network (CNN) techniques.  However, this one-size-fits-all approach has significant downsides: difficulty in modeling long-range interactions, constrained field of view, and difficulty in representing continuous physical states.
As an alternative, the {\em polyline} approach describes curves (\egc lanes, crosswalks, boundaries) as piecewise linear segments.  This is a significantly more compact form due to the sparse nature of road networks. Previous works typically process a set-of-polylines description of the world in a per-agent, agent-centric coordinate system.  LaneGCN~\cite{liang2020laneGCN} stands apart by treating road lanes as nodes in a graph neural network, leveraging road network connectivity structure.

To model \textbf{motion history}, one popular choice is to encode the sequence of past observed states via a recurrent net (GRU, LSTM) or temporal (1D) convolution.  As an alternative, in the {\em raster} framework, the state sequence is typically rendered as a stack of binary mask images depicting agent oriented bounding boxes, or rendered in the same image, with the corresponding time information rendered separately~\cite{Refaat2019AgentPF}.  

To model \textbf{agent interactions}, one must deal with a dynamic set of neighboring agents around each modeled agent.  This is typically done by aggregating neighbor motion history with a permutation-invariant set operator: pooling or soft attention. Notably, Precog~\cite{precog_Rhinehart_2019_ICCV} jointly rolls out agent policies in a step-wise simulation.  Raster approaches rely on convolution over the 2D spatial grid to implicitly capture interactions; long-term interactions are dependent on the network receptive fields.  

Agent \textbf{trajectory decoding} choices are similar to choices for encoding motion history, with the exception of methods that do lookup on a fixed or learned trajectory database~\cite{phan2019covernet,biktairov2020prank}.

The most \textbf{popular output trajectory representation}
is a sequence of {\em states} (or state differences).  
A few works~\cite{rhinehart2018r2p2,rhinehart2019precog} 
instead model Newton's laws of {\em motion} in a discrete time-step aggregation capturing Verlet integration.  
Other works~\cite{salzmann2020trajectron++,cui2020deep} explicitly model {\em controls} which parameterize a kinematically-feasible model for vehicles and bicycles.  With any of these representations, the spacetime trajectory can be intrinsically represented as a sequence of sample points or a continuous polynomial representation~\cite{buhet2020plop}.
In our experimental results, we explore the effectiveness of states and kinematic controls, with and without an underlying polynomial basis.  Notably unique are (1) HOME~\cite{gilles21home} and GOHOME~\cite{gilles2021gohome} which first predict a heatmap, and then decode trajectories after sampling, and (2) MP3~\cite{casas2021mp3} and NMP~\cite{neural_motion_planner_zeng2019} which learn a cost function evaluator of trajectories, and the trajectories are enumerated heuristically rather than generated by a learned model.

Nearly all work assumes an independent, per-agent output space, in which agent interactions cannot be explicitly captured.  A few works are notable in describing joint interactions as output, either in an asymmetric~\cite{wimp2020,tolstaya2021cbp} or symmetric way~\cite{ettinger2021womd,precog_Rhinehart_2019_ICCV,ngiam21scene_transformer}.

The choice of output \textbf{trajectory distribution} has ramifications on downstream applications. An intrinsic property of the driving setting is that a vehicle or a pedestrian can follow one of a diverse set of possible trajectories. It is thus essential to capture the multimodal nature of the problem.  Gaussian
Mixture Models ({\em GMM}s) are a popular choice for this purpose due to their compact parameterized form;
mode collapse is addressed through training tricks~\cite{kitani_diverse_forecasting_dpps, cui2019multimodal} or the use of trajectory anchors~\cite{sapp2019multipath}.  Other approaches model a discrete distribution over a set of trajectories (learned or fixed a priori)~\cite{zhao2020tnt, liang2020laneGCN,biktairov2020prank,cui2019multimodal}, or via a collection of trajectory samples drawn from a latent distribution and decoded by the model~\cite{sociallstm,DESIRE,precog_Rhinehart_2019_ICCV,rhinehart2018r2p2,marchetti2020mantra}.

\section{Model Architecture}
\label{sec:model}
\begin{figure*}[h]
    \centering
    \includegraphics[width=\textwidth,clip=true,trim=0cm 5cm 2cm 3cm]{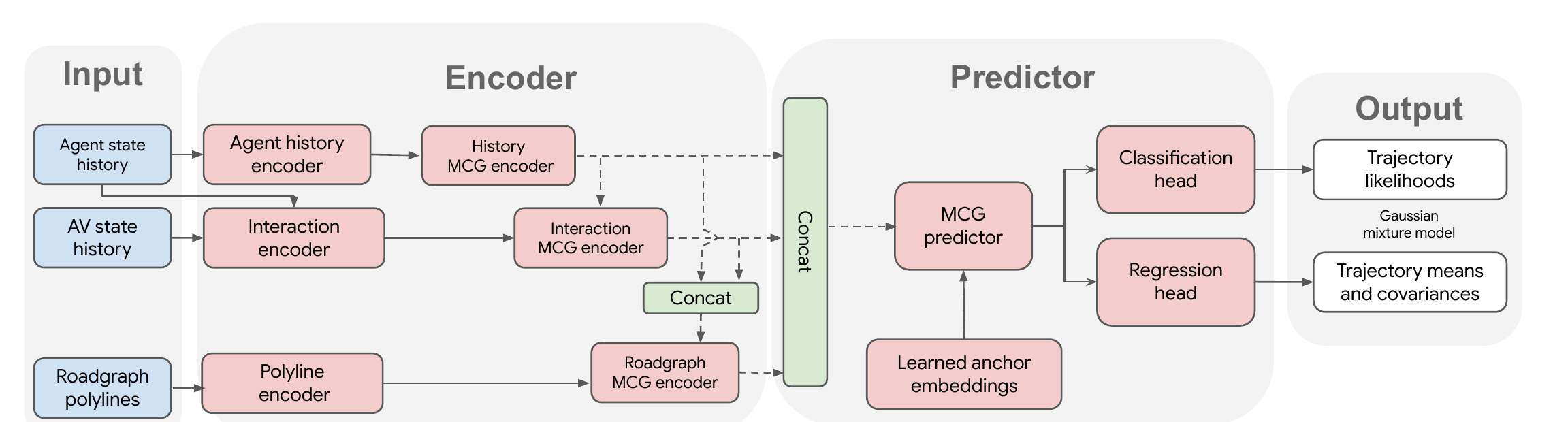}
    \caption{\modelname Model Architecture. MCG denotes Multi-Context Gating, described in Section~\ref{sec:model}. Blocks in red highlight portions of the model with learned parameters. Dotted inputs to the MCG denotes context features. Each of the encoder MCG outputs aggregated embeddings (one per agent) as shown by dotted arrows. On the other hand, the predictor MCG outputs one embedding per trajectory per agent}
    \label{fig:model}
\end{figure*}

Figure \ref{fig:model} depicts the proposed \modelname model architecture, which on a high level is similar to that of MultiPath~\cite{sapp2019multipath};
the model consists of an encoding step and a predictor head which conditions on anchors and outputs a Gaussian Mixture Model (GMM)~\cite{bishop2006pattern} distribution for the possible agent position at each future time step.  

MultiPath used a common, top-down image based representation for all input modalities (e.g., agents' tracked state, road network information), and a CNN encoder.  In contrast, \modelname has encoders processing each input modality and converting it to a compact and sparse representation; the different modality encodings are later fused using a multi-context gating (MCG) mechanism. 


\subsection{Input Representation}
\label{sec:features}
\modelname makes predictions based on the following input modalities:
\begin{itemize}
    \item \emph{Agent state history}: 
    a state sequence describing the agent trajectory for a fixed number of past steps. In the Waymo Open Motion dataset~\cite{ettinger2021womd}, this state information includes position, velocity, 3D bounding box size, heading angle and object type; for Argoverse~\cite{chang2019argoverse} only position information is provided. The state is transformed to an agent-centric coordinate system, such that the most recent agent pose is located at the origin and heading east. \footnote{Since the explicit heading is missing in Argoverse data, we use the last two time steps to get the current orientation.}
    
    \item \emph{Road network}:
    Road network elements such as lane lines, cross walks, and stop lines are often represented as parametric curves like clothoids~\cite{neural_motion_planner_zeng2019}, which can be sampled to produce point collections that are easily stored in multi-dimensional array format, as is done in many public datasets~\cite{ettinger2021womd, chang2019argoverse}.  We further summarize this information by approximating point sequences for each road element as a set of piecewise linear segments, or polylines, similar to~\cite{gao2020vectornet, liang2020laneGCN, homayounfar2018maps}.  
    
    \item \emph{Agent interactions}: 
    For each modeled agent, we consider all neighboring agents. For each neighboring agent, we extract features in the modeled agent's coordinate frame, such as relative orientation, distance, history and speed.
    
    \item \emph{AV-relative features}: Similar to the interaction features, we extract features of the autonomous vehicle / sensing vehicle (AV) relative to each other agent. We model the AV separately from the other agents.  We hypothesize this is a helpful distinction for the model because:
    (a) The AV is the center of sensors' field of view. Tracking errors due to distance and occlusion are relative to this center.
    (b) The behavior of the AV can be unlike the other road users, which to a good approximation can be assumed to all be humans. 
\end{itemize}

Details on how these features are encoded and fused are described next. These steps comprise the ``Encoder'' block of Figure~\ref{fig:model}, whose output is an encoding per agent, in each agent's coordinate frame.

\subsection{Multi~Context Gating for fusing modalities}
\label{sec:mcg}

In this section we focus on how to combine the different input modality encodings in an effective way.  Other works use a common rasterized format~\cite{sapp2019multipath, neural_motion_planner_zeng2019}, a simple concatenation of encodings~\cite{DESIRE,precog_Rhinehart_2019_ICCV, salzmann2020trajectron++}, or employ attention~\cite{ngiam21scene_transformer,tang_multifuture,gao2020vectornet,liang2020laneGCN}. We propose an efficient mechanism for fusing information we term {\em multi-context gating} (MCG), and use MCG blocks throughout the \modelname architecture.

Given a set of elements $\mathbf{s}_{1:N}$ and an input context vector $\mathbf{c}$, a CG block assigns an output $\mathbf{s'}_{1:N}$ to each element in the set, and computes an output context vector $\mathbf{c'}$.
The output does not depend on the ordering of input elements. Mathematically, let $\cg(\cdot, \cdot)$ be the function implemented by the CG block, and $\pi$ be any permutation operation on a sequence of $n$ elements.
The following equations hold for CG:
\begin{equation}
\begin{array}{ll}
     (\mathbf{s}'_{1:n}, \mathbf{c}') & = \cg(\mathbf{s}_{1:n}, \mathbf{c})  \\
     (\mathbf{s}^*_{1:n}, \mathbf{c}^*) & = \cg(\pi(\mathbf{s}_{1:n}), \mathbf{c}) \\
\end{array}    
\end{equation}
which imply that we have
\begin{equation*}
\begin{array}{lll}
     \mathbf{c}^* &= \mathbf{c}' & \text{(permutation-invariance)}\\
     \mathbf{s}^*_{1:n} &= \pi(\mathbf{s}'_{1:n}) & \text{(permutation-equivariance).}\\
\end{array}    
\end{equation*}
The size of the set $n$ can vary across calls to $\cg(\cdot, \cdot)$. 

CG's set function properties---permutation invariance/equivariance and ability to process arbitrarily sized sets---are naturally motivated by the need to encode a variable, unordered set of road network elements and agent relationships. 
A number of set functions have been proposed in the literature such as 
DeepSets~\cite{zaheer17deepset}, PointNet~\cite{qi2017pointnet} and SetTransformers~\cite{lee19settransformer}.

\begin{figure}
    \centering
    \includegraphics[width=0.95\textwidth]{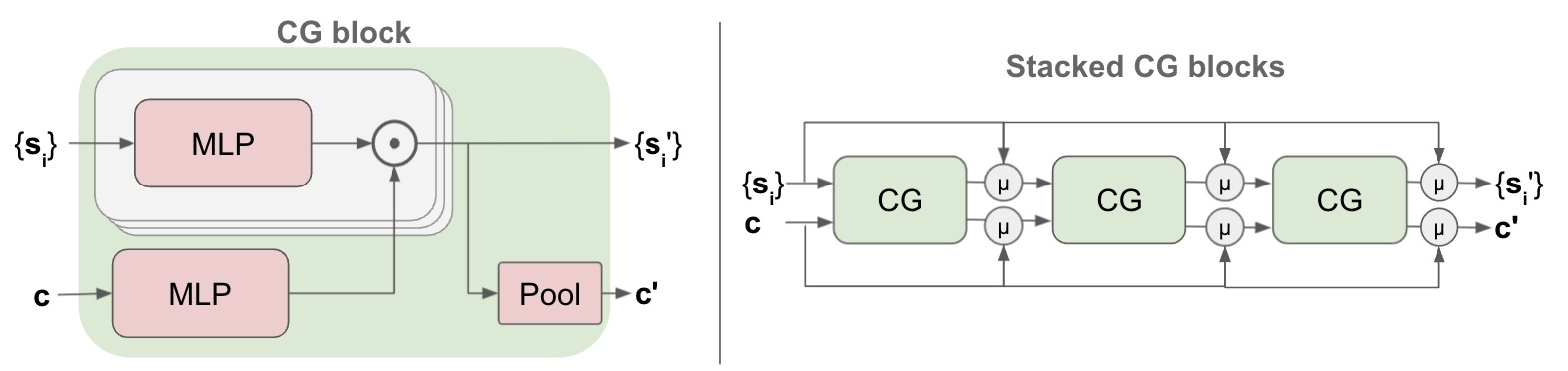}
    \caption{\textbf{Left:} Context gating (CG) block diagram. \textbf{Right:} 3 CG blocks stacked together, with running-average skip-connections (shown as components labeled ``$\mu$'').  See Section~\ref{sec:mcg} for details.}
    \label{fig:mcg}
\end{figure}

A single CG block is implemented via
\begin{eqnarray}
\tilde{\mathbf{s}}_i & = & \operatorname{MLP}(\mathbf{s}_i) \\
\tilde{\mathbf{c}} & = & \operatorname{MLP}(\mathbf{c}) \\
\mathbf{s}'_{i} & = & \tilde{\mathbf{s}}_i \odot \tilde{\mathbf{c}} \\
\mathbf{c}' & = & \operatorname{Pool}(\mathbf{s}'_{1:n}),
\end{eqnarray}
where $\odot$ denotes element-wise product and $\operatorname{Pool}$ is a permutation-invariant pooling layer 
such as max or average pooling.  These operations are illustrated in Figure~\ref{fig:mcg}.
In the absence of an input context, we simply set $\tilde{\mathbf{c}}$ to an all-ones vector in the first context gating block.
Note that both $\mathbf{s}'_{i}$ and $\mathbf{c}'$ depend on all inputs.
It can be shown that $\mathbf{c}'$ is permutation-invariant w.r.t the input embeddings.
It can also be shown that $\mathbf{s}'_{i}$ are permutation-equivariant.

We stack multiple CG blocks by incorporating running-average skip-connections, as is done residual networks~\cite{ResNet16}:
\begin{eqnarray}
\mathbf{\bar{s}}^k_{1:n} & =\frac{1}{k} \sum_{j=1}^k \mathbf{s}^j_{1:n} \\
\mathbf{\bar{c}}^k & =\frac{1}{k} \sum_{j=1}^k \mathbf{c}^j \\
\left(\mathbf{s}^{k+1}_{1:n}, \mathbf{c}^{k+1}\right) & = \cg\left(\mathbf{\bar{s}}_{1:n}^{k}, \mathbf{\bar{c}}^k \right).
\end{eqnarray}

We denote such multi-layer CG blocks as $\mcg_N(\cdot, \cdot)$ for a stack of $N$ $\cg$ blocks.


\textbf{Comparison with attention.} Attention is a popular mechanism in domains such as NLP~\cite{vaswani2017attention} and computer vision~\cite{dosovitskiy2020ViT,dai2021coatnet}, in which the encoding for each element of a set is updated via a combination of encodings of all other elements.  For a set of size $n$, this intrinsically requires $O(n^2)$ operations.  In models of human behavior in driving scenarios, {\em self} attention has been employed to update encodings for, \eg, road lanes, by attending to neighboring lanes, or to update encodings per agent based on the other agents in the scene.  {\em Cross} attention has also been used to condition one input type (\eg agent encodings) on another (\eg road lanes)~\cite{liang2020laneGCN, gao2020vectornet, ngiam21scene_transformer}.  Without loss of generality, if there are $n$ agents and $m$ road elements, this cross attention scales as $O(nm)$ to aggregate road information for each agent.

\begin{figure*}
    \centering
    \includegraphics[width=0.5\textwidth]{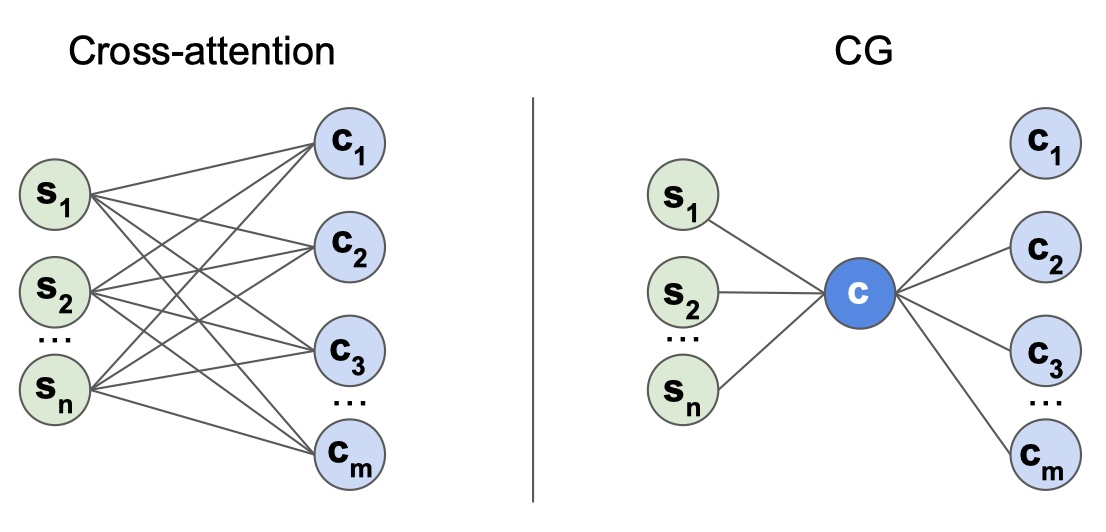}
    \caption{A comparison of the {\em element relationship graph} for cross-attention and CG.  In cross-attention, each element $\mathbf{s}_i$ aggregates information from $\mathbf{c}_{1:m}$.  In CG, $\mathbf{c}_{1:m}$ summarized with a single context vector $\textbf{c}$.} 
    \label{fig:mcg-relationships}
\end{figure*}

$\cg$ can be viewed as an approximation to cross-attention.   Rather than each of $n$ elements attending to all $m$ elements of the latter set, CG summarizes the latter set with the {\em single} context vector $\mathbf{c}$, as shown in Figure~\ref{fig:mcg-relationships}. Thus the dimensionality of $\mathbf{c}$ needs to be great enough to capture all the useful information contained in the original $m$ encodings. If the dimensionality of elements is $d$, and the dimensionality of $\mathbf{c}$ is $d^c$, then if $d^c = m d$, CG can be reduced to some form of cross-attention by setting $\mathbf{c} = \operatorname{Concat}(\{c_j\}_{j=1}^m)$.  When $d^c < md$, we are trading off the representational power of full cross-attention with computational efficiency.

\subsection{Encoders}

In this section we detail the specific encoders shown in Figure~\ref{fig:model}.

\textbf{Agent history encoding. } The agent history encoding is  obtained by concatenating the output of three sources: 

\begin{enumerate}
    \item A LSTM on the history features from $H$ time steps ago to the present time: $(\mathbf{x}_t, \mathbf{y}_t)_{t=-H:0}$.
    \item A LSTM on the {\em difference} in the history features  $(\mathbf{x}_t-\mathbf{x}_{t-1},\mathbf{y}_t-\mathbf{y}_{t-1})_{t=-H+1:0}$.
    \item MCG blocks applied to the set of history elements.  Each element in the set consists of a historical position and time offset in seconds relative to the present time. The context input here is an all-ones vector with an identity context MLP. Additionally we also encode the history frame id as a one hot vector to further disambiguate the history steps.
\end{enumerate}

We denote the final embedding, which concatenates these three state history encodings, as $\phi^\mathrm{(state)}$.

\textbf{Agent interaction encoding}.  For each modeled agent, we build an interaction encoding by considering each neighboring agent $\nu$'s past state observations: $\{\bfx_{-H}^\nu,\ldots, \bfx_{-1}^\nu, \bfx_{0}^\nu\}$.  We transform $\nu$'s state into the modeled agent's coordinate frame, and embed it with a LSTM to obtain an embedding $\phi^{(\mathrm{interaction})}_{\nu}$. 
Note this is similar to the ego-agent history embedding but instead applied
to the \emph{relative coordinates of another agent}.

By doing this for $n$ neighboring agents we obtain a set of interaction embeddings $\phi^{(\mathrm{interaction})}_{i=1:n}$.  We fuse neighbor information with stacked MCG blocks as follows

\begin{equation}
(\phi'^{(\mathrm{interaction})}_{1:n}, \phi'^{\mathrm{(state)}}) = \mcg_N(\phi^{(\mathrm{interaction})}_{1:n},  [\phi^\mathrm{(state)}, \phi^{(\mathrm{interaction})}_\mathrm{AV}])
\end{equation}
where the second argument is the input context vector to $\mcg$, which in this case is a concatenation of the modeled agent's history embedding, and the AV's interaction embedding. In this way we emphasize the AV's representation as a unique entity in the context for all interactions; see Section~\ref{sec:features} for motivation.

\textbf{Road network encoding}. We use the polyline road element representation discussed in Section~\ref{sec:features} as input. 
Each line segment is parameterized by its start point, end point and the road element semantic type $\Upsilon$ (\eg, \emph{ Crosswalk}, \emph{SolidDoubleYellow}, \etc). For each agent of interest, we transform the closest $P=128$ polylines into their frame of reference and call the transformed segment $\bfp = (\bfa, \bfb)$. Let $\bfr$ be the closest point from the agent to the segment, and $\bfa^\perp$ be the unit tangent vector at $\bfa$ on the original curve.
Then we represent the agent's spatial relationship to the segment via the vector 
$[||\bfr||_2, \bfr/||\bfr||_2, (\bfb-\bfa)/||\bfb-\bfa||_2, ||\bfb-\bfa||_2, ||\bfb-\bfr||_2, \bfa^\perp, \operatorname{OneHotEncoding}(\Upsilon)]$.  These feature vectors are each processed with a shared MLP,
resulting in a set of agent-specific embeddings per road segment, which we denote by $\phi^{(\mathrm{road})}_{1:P}$.
We then fuse road element embeddings
with the agent history embedding $\phi^{\mathrm{(state)}}$ using stacked MCG blocks
\begin{equation}
    (\phi'^{(\mathrm{road})}_{1:P}, \phi'^{\mathrm{(state)}}) = \mcg_N(\phi^{(\mathrm{road})}_{1:P}, \phi^{\mathrm{(state)}})
\end{equation}
and thus enrich the road embeddings with dynamic state information.

\subsection{Output representation}
\label{sec:output}
\modelname predicts a distribution of future behavior parameterized as a Gaussian Mixture Model (GMM), as is done in MultiPath~\cite{sapp2019multipath} and other works~\cite{mercat2020multi,phan2019covernet,buhet20_plop}.  For efficient long-term prediction, the distribution is conditionally independent over time steps across mixture components, thus each mode at each time step is represented as a Gaussian distribution over $(x,y)$ with a mean $\mu^t \in \mathbb{R}^2$ and covariance $\Sigma^t \in \mathbb{R}^{2 \times 2}$.  The $M$ mode likelihoods $p_{1:M}$ are tied over time. MAP inference per mode is equivalent to taking the sequence $\mu^{1:T}$ as state waypoints defining a possible future trajectory for the agent.  The full output distribution is
\begin{equation}
p(s) = \sum_{i=1}^M p_i \prod_{t=1}^T \mathcal{N}(s_t - \mu_i^t, \Sigma_i^t)
\end{equation}
where $s = s^{1:T}$ represents a trajectory; $s^t \in \mathbb{R}^2$.

The classification head of Figure~\ref{fig:model} predicts the $p_i$ as a softmax distribution over mixture components.  The regression head outputs the parameters of the Gaussians $\mu$ and $\Sigma$ for $M$ modes and $T$ time steps.

\textbf{Training objective.} We follow the original MultiPath approach and maximize the likelihood of the groundtruth trajectory under our model's predicted distribution.  We make a hard-assignment labeling of a ``correct'' mixture component by choosing the one with the smallest Euclidean distance to the groundtruth trajectory.

The average log loss over the entire training set is optimized using Adam. We use an initial learning rate of $0.0002$ and a batch size of $64$, with decay rate of $0.5$ every 2 epochs. The final model is chosen after training for $800,000$ steps.

\subsection{Prediction architecture with learned anchor embeddings}
The goal of the Predictor module (Figure~\ref{fig:model}) is to predict the parameters of the GMM described in Section~\ref{sec:output}, namely $M$ trajectories, with likelihoods and uncertainties around each waypoint.

In applications related to future prediction, capturing the highly uncertain and multimodal set of outcomes is a key challenge and the focus of much work~\cite{rhinehart2018r2p2, liang2020garden, kitani_diverse_forecasting_dpps, phan2019covernet, sapp2019multipath}.  One of MultiPath's key innovations was to use a static set of {\em anchor trajectories} as pre-defined modes that applied to all scenes. One major downside to this is that most modes are not a good fit to any particular scene, thus requiring a large amount modes to be considered, with most obtaining a low-likelihood and getting discarded.  Another downside is the added complexity and effort stemming from a 2-phase learning process (first estimating the modes from data, then training the network).

In this work, we learn anchor embeddings as part of the overall model training. We interpret these embeddings $\mathbf{e}_{1:M}$ as anchors in latent space, and construct our architecture to have a one-to-one correspondence with these embeddings and the output trajectory modes of our GMM.
The vectors $\mathbf{e}_{1:M}$
are trainable model parameters that are independent of the input.  This has connections to Detection Transformers (DETR)~\cite{carion20detr} which propose a way to learn anchors rather than hand-design them for object detection.  This is also similar in spirit to MANTRA~\cite{marchetti2020mantra}, a trajectory prediction network, which has an explicit learned memory network which consists of a database of embeddings that can be retrieved and decoded into trajectories.

We concatenate the embeddings $\phi'^{\mathrm{(interaction)}}$, $\phi'^{\mathrm{(road)}}$ and $\phi'^{\mathrm{(state)}}$ obtained from the output of the respective $\mcg$ blocks to obtain a fixed-length feature vector $\phi^{\mathrm{(combined)}}$ for each modeled agent.  We then use this as context in stacked MCG blocks that operate on the set of anchor embeddings $\mathbf{e}_{1:M}$, with a final MLP that predicts all parameters of the output GMM:

\begin{displaymath}
(\mu, \sigma_x, \sigma_y, \rho_{xy}, \log q)_{1:M} = \operatorname{MLP}(\mcg(\mathbf{e}_{1:M},\phi^{\mathrm{(combined)}})),
\end{displaymath}
where $\Sigma$ is formed from $(\sigma_x, \sigma_y, \rho_{xy})$.

\subsection{Internal Trajectory Representation}
\label{sec:traj_decoding}
We model the future position and heading of agents, along with agent-relative longitudinal and lateral Gaussian uncertainties.
We parameterize the trajectory using $x(t), y(t), \theta(t), \sigma_{lng}(t), \sigma_{lat}(t)$---position, heading, and standard deviation for longitudinal and lateral uncertainty.

The most popular approach in the literature is to directly predict a sequence of such states at uniform time-discretization.  Here we also consider two non-mutually exclusive variants.

\begin{enumerate}

\item We can represent functions over time as polynomials, which add an inductive bias that ensures a smooth trajectory. It gives us a compact, interpretable representation of each predicted signal.

\item Instead of directly predicting $\{x(t), y(t), \theta(t)\}$, we can predict the underlying kinematic control signals, which can then be integrated to evaluate the output state.
In this work, we experiment with predicting the acceleration $a(t)$ and heading change rate $\dot{\theta}(t)$ and integrating them to recover the trajectory as follows:
\begin{eqnarray*}
\begin{array}{cl}
    v(t) & = v(0) + \int_0^t a(\tau) d\tau  \\
     \theta(t) & = \theta(0) + \int_0^t \dot{\theta}(\tau) d\tau \\
     x(t) & = x(0) + \int_0^t v(\tau) \cos(\theta(\tau)) d\tau \\
     y(t) & = y(0) + \int_0^t v(\tau) \sin(\theta(\tau)) d\tau.
\end{array}
\label{eq:control_integration}
\end{eqnarray*}

\end{enumerate}

These representations add inductive bias encouraging natural and realistic trajectories that are based on realistic kinematics and consistent with the current state of the predicted agent. For the polynomial representation, it is also possible to specify a \emph{soft constraint} by regularizing the polynomial's constant term, which determines the shift of the predicted signal from its current value.

Algorithm \ref{alg:ctrldecoder} demonstrates the conversion from control signals to output positions.
Note that this operation is differentiable, permitting end-to-end optimization. It is a numerical approximation of Equation~\ref{eq:control_integration} with additional 
technical considerations: (1) When computing the next position $(x(t), y(t))$, we use the midpoint approximation of the speed and heading
$\tilde{\theta}(t) \equiv \theta(t-\delta_t) + \dot{\theta}(t-\delta_t)\cdot(\delta_t/2)$.  (2) Given vehicle dimensions, we cap the heading change rate to match a predetermined maximum feasible curvature.
(3) These equations are applied to the rear-axle of the vehicle rather than the center position.
We use the rear-end position of the vehicle as an approximation of the rear-axle position.

\RestyleAlgo{boxed}
\begin{algorithm}
\label{alg:ctrldecoder}
{\small
\KwData{$a(t)$ and $\dot{\theta}(t)$ at $t=0,0.1,0.2,\dots,10.0$, $x(0), y(0), \theta(0), v(0)$}
\KwResult{$x(t)$, $y(t)$ and $\theta(t)$ at $t=0.1,0.2,\dots,10.0$}
$\delta_t = 0.1$ \\
\For{$t = 0.1,0.2,\dots,10.0$}{
$\tilde{v} \leftarrow v(t-\delta_t) + a(t) \cdot (\delta_t/2)$ \\
$\dot{\theta}_{\rm cap} \leftarrow {\rm CapCurvature}(\dot{\theta}(t-\delta_t))$ \\
$\tilde{\theta} \leftarrow \theta(t-\delta_t) + \dot{\theta}_{\rm cap}\cdot(\delta_t/2)$ \\
$x(t) \leftarrow x(t-\delta_t) + \tilde{v} \cos(\tilde{\theta})\delta_t$ \\
$y(t) \leftarrow y(t-\delta_t) + \tilde{v} \sin(\tilde{\theta})\delta_t$ \\
$\theta(t) \leftarrow \theta(t-1) + \dot{\theta}_{\rm cap}\delta_t$ \\
$v(t) \leftarrow \theta(t-1) + a(t-\delta_t)\delta_t$
}
\caption{Integrating control-signal to positions.}
}
\end{algorithm}

Note that Algorithm \ref{alg:ctrldecoder} can be viewed as a special type of recurrent network, without learned parameters.  This decoding stage then mirrors other works which use a learned RNN (LSTM or GRU cells) to decode an embedding vector into a trajectory~\cite{mercat2020multi,wimp2020,hong2019rules,tang_multifuture,salzmann2020trajectron++}. In our case, the recurrent network state consists of $x(t), y(t), v(t)$ and $\theta(t)$, and the input consists of $\dot{\theta}(t)$ and $a(t)$. 
Encoding an inductive bias  derived from kinematic modeling spares the network the need to explicitly learn these properties makes the predicted state compact. This promotes data efficiency and generalization power, but can be more sensitive to perception errors in the current state estimate. 

\section{Ensembling predictor heads via bootstrap aggregation}
\label{sec:agg}

Ensembling is a powerful and popular technique in many machine learning applications. For example, ensembling is a critical technique for getting the best performance on ImageNet~\cite{ResNet16}.  By combining multiple models which are to some degree complementary, we can enjoy the benefits of a higher capacity model with lower statistical variance.

We specifically apply bootstrap aggregation (bagging)~\cite{eslbook} to our predictor heads by training $E$ such heads together.  To encourage models learning complementary information, the weights of the $E$ heads are initialized randomly, and an example is used to update the weights of each head with a 50\% probability.

Unlike scalar regression or classification, it is not obvious how to combine output from different heads in our case---each is a Gaussian Mixture Model, with no correspondence of mixture components across ensemble heads.  Furthermore, we consider allowing each predictor head to predict a richer output distribution with more modes $L > M$; where $M$ is fixed as a requirement for the task (and is used in benchmark metrics calculations).

Let $\operatorname{\Psi}$ denote the union of the predictions
from all heads
\begin{equation}
\operatorname{\Psi} = \{ (\mathbf{\mu}_1, \mathbf{\Sigma}_1, q_1), \ldots, (\mathbf{\mu}_{M'}, \mathbf{\Sigma}_{M'}, q_{M'}) \},
\end{equation}
where $M' = L \cdot E$, and the mode likelihoods $p$ are divided by the number of heads $E$ so that they sum up to 1. Then we pose the ensemble combination task as one of converting
$\operatorname{\Psi}$ to a more compact GMM $\bar{\operatorname{\Psi}}$ with $M$ modes:
\begin{equation}
\bar{\operatorname{\Psi}} = \left( (\bar{\mathbf{\mu}}_1, \bar{\mathbf{\Sigma}}_1, \bar{q}_1), \ldots, (\bar{\mathbf{\mu}}_M, \bar{\mathbf{\Sigma}}_M, \bar{q}_M) \right),
\end{equation}
while requiring that  $\bar{\operatorname{\Psi}}$ best approximates $\operatorname{\Psi}$.
In this section we describe the aggregation algorithm we use. Theoretical motivations and derivation can be found in Appendix~\ref{sec:appendix-aggregation}.

We find fit $\bar{\Psi}$ to $\Psi$ using an iterative clustering algorithm, like Expectation-Maximization~\cite{bishop2006pattern}, but with hard assignment of cluster membership.  This setting lends itself to efficient implementation in a compute graph, and allows us to train this step end-to-end as a final layer in our deep network.

We start by selecting $M$ cluster centroids from $\mu_{1:M'}$ in a greedy fashion.
The selection criteria is to maximize the
probability that a centroid sampled from $\operatorname{\Psi}$ lies within $\tau$ distance from at least one selected centroid:
\begin{equation}
  \mathbf{\bar{\mu}}_{1:M} = \underset{\mathbf{\bar{\mu}}_{1:M}}{\operatorname{argmax}} \sum_{i=1}^{M'} q_i\,\, \underset{{\bar{\mu} \in   \mathbf{\bar{\mu}}_{1:M}}}{\operatorname{max}}\,\, \mathbb{I} \left(\Vert \mathbf{\mu}_i -\bar{\mu} \Vert_2 \leq \tau \right )
\label{eq:em_init_obj}
\end{equation}
This is a criterion that explicitly optimizes trajectory diversity, which is a fit for metrics such as miss rate, mAP and minADE, as defined in ~\cite{chang2019argoverse, ettinger2021womd}. Other criteria could also be used depending on the metric of interest.  
It is interesting to relate this criteria to the ensembling and sampling method employed by GOHOME~\cite{gilles2021gohome}.  In that work, they output an intermediate spatial heatmap representation, which is amenable to ensemble aggregation.  Then they greedily sample end-points in a similar fashion.

Since jointly optimizing $\mathbf{\bar{\mu}}_{1:M}$ is hard, we select each $\mu_i$ greedily for $i=1,\ldots,M$ according to
\begin{equation}
  \mathbf{\bar{\mu}}_i = \underset{\mathbf{\bar{\mu}}_i}{\operatorname{argmax}} \sum_{i=1}^{M'} q_i\,\, \underset{{\bar{\mu} \in   \mathbf{\bar{\mu}}_{1:i}}}{\operatorname{max}}\,\, \mathbb{I} \left(\Vert \mathbf{\mu}_i -\bar{\mu} \Vert_2 \leq \tau \right )
\label{eq:em_init_greedy}
\end{equation}
which differs in that the outer $\operatorname{argmax}$ is done iteratively over $\bar{\mu}_{i}$ rather than jointly $\bar{\mu}_{1:M}$.

Starting with the selected centroids, We iteratively update the parameters of $\bar{\operatorname{\Psi}}$ using an expectation-maximization-style~\cite{dempster77} algorithm, where each iteration consists of the following updates

\begin{align}
\bar{q}_h &\leftarrow \sum_{i=1}^{M'} q_i p(h | \mathbf{\mu}_i; \bar{\Psi}) \\   
\bar{\mathbf{\mu}}_h &\leftarrow \frac{1}{\bar{q}_h} \sum_{i=1}^{M'} q_i p(h | \mathbf{\mu}_i; \bar{\Psi})  \mathbf{x} \\   
\bar{\mathbf{\Sigma}}_h &\leftarrow \frac{1}{\bar{q}_h} \sum_{i=1}^{M'} q_i  p(h | \mathbf{\mu}_i; \bar{\Psi}) \left[ \mathbf{\Sigma}_i +  (\mathbf{\mu}_i - \bar{\mathbf{\mu}}_h) (\mathbf{\mu}_i - \bar{\mathbf{\mu}}_h)^T \right],  
\end{align}

where $p(h | \mathbf{x}; \bar{\Psi})$ is the posterior probability 
that a given sample $\mathbf{x}$ is sampled from the $h^{th}$ component
of the mixture model specified by $\bar{\Psi}$, which can be computed as
\begin{equation}
p(h | \mathbf{x}; \bar{\Psi}) = \frac{\bar{q}_h\mathcal{N}(\mathbf{x}- \mathbf{\bar{\mu}}_h, \mathbf{\bar{\Sigma}}_h) }{\sum_{k=1}^M \bar{q}_k\mathcal{N}(\mathbf{x}- \mathbf{\bar{\mu}}_k, \mathbf{\bar{\Sigma}}_k)}
\end{equation}

\section{Experiments}
\subsection{Datasets}
The Waymo Open Motion Dataset (WOMD)~\cite{ettinger2021womd} consists of 1.1M examples time-windowed from 103K 20s scenarios. The dataset is derived from real-world driving in urban and suburban environments. Each example for training and inference consists of 1 second of history state and 8 seconds of future, which we resample at 5Hz.  The object-agent state contains attributes such as position, agent dimensions, velocity and acceleration vectors, orientation, angular velocity, and turn signal state. 
The long (8s) time horizon in this dataset tests the model's ability to capture a large field of view and scale to an output space of trajectories, which in theory grows exponentially with time. 

The Argoverse dataset~\cite{chang2019argoverse} consists of 333K scenarios containing trajectory histories, context agents, and lane centerline inputs for motion prediction. The trajectories are sampled at 10Hz, with 2 seconds of past history and a 3-second future prediction horizon.


\subsection{Metrics}
We compare models using competition specific metrics associated with various datasets\footnote{
For each dataset, we report the results of our model against published
results of publicly available models.
},

Specifically, we report the following metrics. 

\mypar{minDE$_k^t$ (Minimum Distance Error):} The minimum distance, over the top k most-likely trajectories, between a predicted trajectory and the ground truth trajectory at time $t$.

\mypar{minADE$_k$ (Minimum Average Distance Error):} Similar to minDE$_k^t$, but the distance is calculated as an average over all timesteps.

\mypar{MR$_k^t$@$d$ (Miss Rate):} Measures the rate at which minFDE$_k^t$ exceeds $d$ meters.  Note that WOMD leaderboard uses a different definition~\cite{ettinger2021womd}.

\mypar{mAP:} For each set of predicted trajectories, we have at most one positive - the one closest to the ground truth and which is within $\tau$ distance from the ground truth. The other predicted trajectories are reported as misses. From this, we can compute precision and recall at various thresholds. Following WOMD metrics definition~\cite{ettinger2021womd} the agents future trajectories are partitioned into behavior buckets, and an area under the precision-recall curve is computed using the possible true positive and false positives per agent, giving us Average Precision per behavior bucket.  The total mAP value is a mean over the AP's for each behavior bucket.

\mypar{Overlap rate:} The fraction of times the most likely trajectory prediction of any agent overlaps with a real future trajectory of another agent (see \cite{ettinger2021womd} for details).

\mypar{TRI: (Turning Radius Infeasibility)} We compute the turning radius along the predicted trajectories using two approaches: one that uses the predicted yaw output from the model (\textbf{TRI-h}), and the other that doesn't require yaw predictions and instead uses the circumradius constituting three consecutive waypoints (\textbf{TRI-c}). If the radius is less than a certain threshold $\tau$, it is treated as a violation. We set this threshold as the approximate minimum turning radius threshold for a midsize sedan, $\tau=3.5m$. Note that a model that simply predicts a constant heading can achieve a TRI-h rate of zero, hence we also compute inconsistencies between turning radius suggested by the coordinates and the predicted headings (\textbf{TRI-hc}). TRI-hc inconsistency is true when the difference in heading based on circumradius from waypoints and predicted headings is greater than 0.05 radians at any time step in a trajectory.



\subsection{MultiPath baseline}
\label{sec:og_mp}

As our work evolved from MultiPath, we include a reference MultiPath model where the input and backbone are faithful to the original paper~\cite{sapp2019multipath} for a point of comparison, with a few minor differences. Specifically, we use a top-down rendering of the scene as before, but now employ a {\em splat rendering}~\cite{zwicker2001surface_splat} approach for rasterization, in which we sample points uniformly from scene elements and do an orthographic projection.  This is a simpler, sparse form of rendering, which doesn't employ anti-aliasing, but is efficient and straightforward to implement in TensorFlow and run as part of the model compute graph on hardware accelerators (GPU/TPU).

As in the original paper, we use a grid of $400 \times 400$ cells, with grid cell physical dimension of $0.2m \times 0.2m$, thus a total field-of-view of $80m$ centered around the AV sensing vehicle in WOMD, with a ResNet18 backbone~\cite{ResNet16}. We use 128 static anchors obtained via k-means, which are shared among all agent types (vehicles, pedestrians, cyslists) for simplicity.  Figure~\ref{fig:og_mp} illustrates this model's inputs and architecture.

\subsection{External benchmark results}
On Argoverse, {\modelname} achieves top-5 performance on most metrics (Table~\ref{tbl:argoverse}).
Our technique is ranked $1^{st}$ on all metrics on Waymo Open Motion Dataset~\cite{ettinger2021womd} (Table~\ref{tbl:womd}).

The tested model is based on the best configuration in Table \ref{tab:ablation}, where the outputs from multiple ensemble heads are aggregated as described in Section~\ref{sec:agg}.

On WOMD, we also see that the original MultiPath model, even with the refinement of learned anchors and ensembling, is outperformed by more recent methods.  It is interesting to note that MultiPath is the best performing top-down scene-centric model employing a CNN; every known method which outranks it uses sparse representations.
\begin{table}
\centering
\begin{tabular}{l | c c c c c}
\hline
\hline
\multicolumn{4}{c}{Argoverse leaderboard ($k=6$, $d=2m$, $t=3s$)} \\
\hline
                 & Rank$^\dagger$ & brier-minDE & minFDE  & MR      & minADE   \\ 
\hline
LaneGCN~\cite{liang2020laneGCN}       & 50   & 2.059       & 1.364   & 0.163   & 0.868    \\ 
DenseTNT~\cite{gu21dense_tnt}         & 23   & 1.976       & 1.282   & 0.126   & 0.882    \\
HOME + GOHOME~\cite{gilles2021gohome} & 10   & 1.860       & 1.292   & \bf{0.085}   & 0.890    \\
TPCN++~\cite{ye2021tpcn}              & 5    & 1.796       & \bf{1.168}   & 0.116   & \bf{0.780}    \\
\bf{\modelname (ours)}                       & 4    & 1.793       & 1.214   & 0.132   & 0.790    \\
QCraft Blue Team                      & 1    & \bf{1.757}       & 1.214   & 0.114   & 0.801    \\
\hline
\end{tabular}
\caption{Comparison with select state-of-the-art methods on Argoverse leaderboard. $k$ is the number of trajectories. $d$ is the maximum distance for no miss, and $t$ is the trajectory time duration.
\hspace{\textwidth}  
$^\dagger$ Rank on the public leaderboard \url{https://eval.ai/web/challenges/challenge-page/454/leaderboard/1279} as of November 12, 2021, which is sorted by brier-minDE.}
\label{tbl:argoverse}
\end{table}

\begin{table}
\centering
\begin{tabular}{l|c c c c c c c }
\hline
\hline
\multicolumn{4}{c}{Waymo Open Motion Prediction ($k=6$, $t=8s$)} \\
\hline
 & Rank$^\dagger$ & minDE & minADE & MR & Overlap & mAP \\
\hline
MultiPath~\cite{sapp2019multipath}                     & 11  & 2.04 & 0.880  & 0.345 & 0.166 & 0.409   \\
SceneTransformer~\cite{ngiam21scene_transformer}       & 7 &  1.212 & 0.612 & 0.156 & 0.147 & 0.279 \\
DenseTNT~\cite{gu21dense_tnt}                          & 5 &  1.551 & 1.039 & 0.157 & 0.178 & 0.328 \\
\bf{\modelname}                                        & 1 & \bf{1.158} & \bf{0.556} & \bf{0.134} & \bf{0.131} & \bf{0.409} \\ 
\hline
\end{tabular}
\caption{Comparison with published state-of-the-art methods on WOMD public leaderboard. $k$ is the number of trajectories and $t$ is the trajectory time horizon.
\hspace{\textwidth}
$^\dagger$ Rank on the public leaderboard 
\url{
https://waymo.com/open/challenges/2021/motion-prediction/}
as of November 12, 2021, which is sorted by mAP.}
\label{tbl:womd}
\end{table}

\subsection{Qualitative Examples}
\begin{figure}[p!]
    \centering
    \begin{tabular}{ccc}
    \includegraphics[width=0.3\columnwidth]{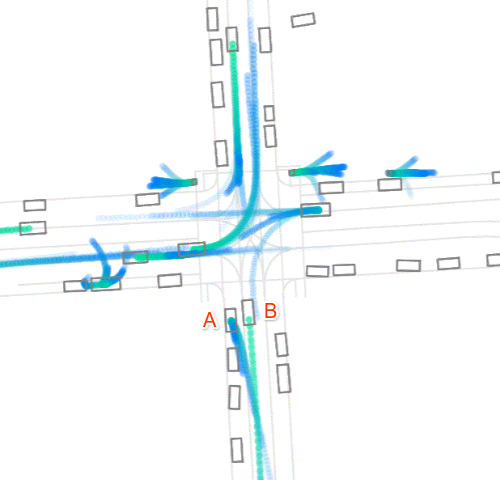} &
    \includegraphics[width=0.3\columnwidth]{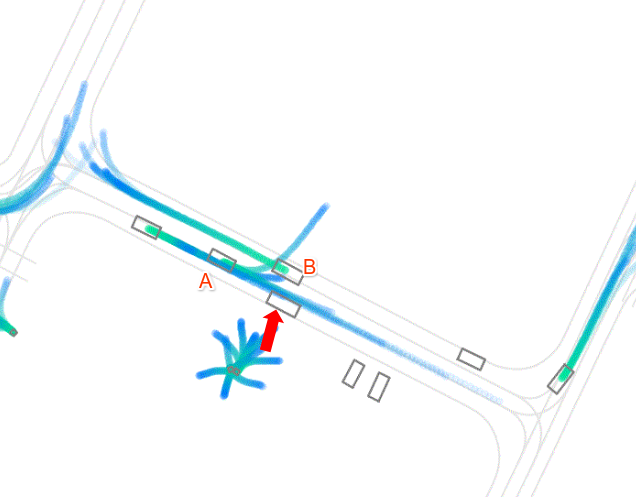} &
    \includegraphics[width=0.3\columnwidth]{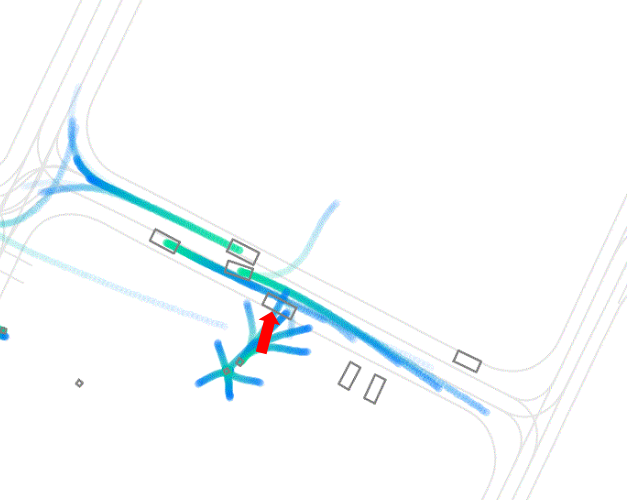} \\
    (a) & (b) & (c) \\ 
    \includegraphics[width=0.3\columnwidth]{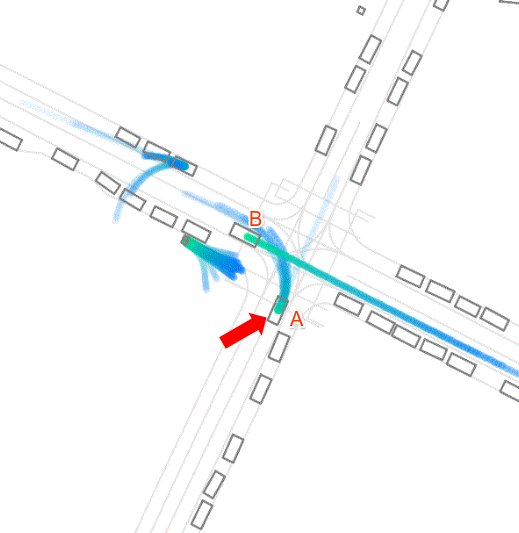} &
    \includegraphics[width=0.3\columnwidth]{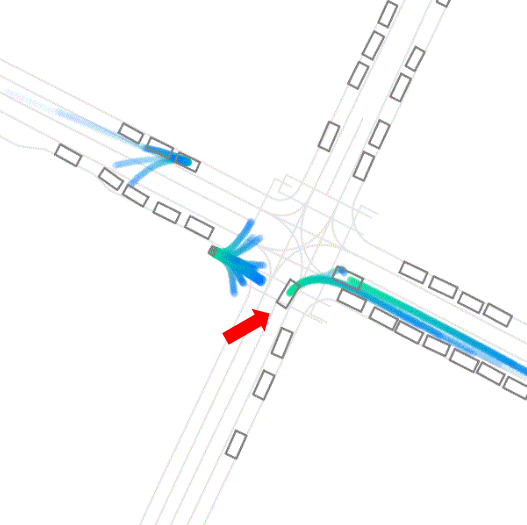} &
    \includegraphics[width=0.3\columnwidth]{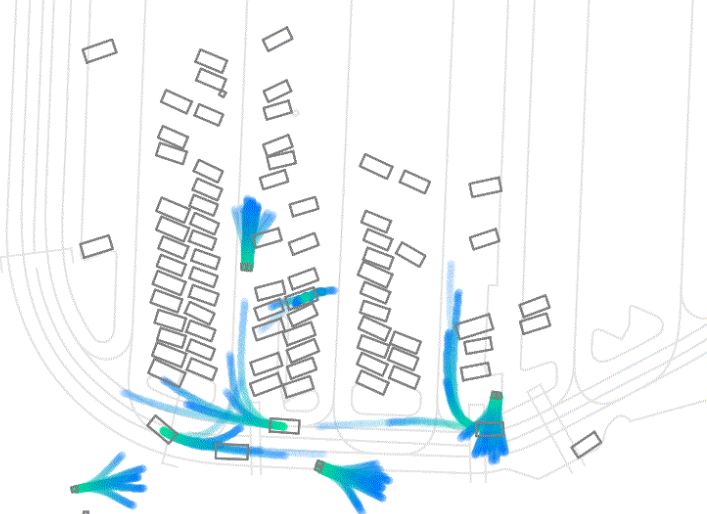} \\
    (d) & (e) & (f) \\
    \includegraphics[width=0.3\columnwidth,clip=true,trim=0cm 1cm 0cm 5cm]{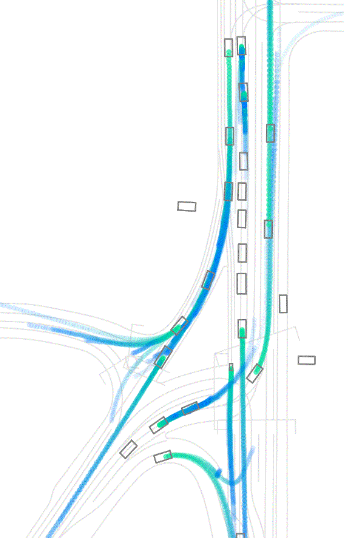} & 
    \multicolumn{2}{c}{\includegraphics[scale=0.4]{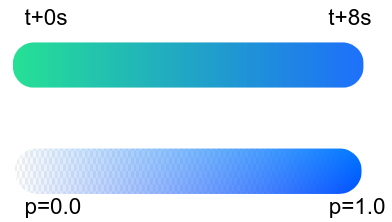}}
    \\
    (g)
    \end{tabular}
    \caption{Examples of \modelname predictions for 8 seconds in WOMD scenes. 
    Hue indicates time horizon while transparency indicates predicted probability. Rectangles indicate vehicles while small squares indicate pedestrians.
    (a): A four-way intersection involving multiple interactions. For example, car A is predicted to yield for car B. (b \& c): Narrow road interaction. Car A is predicted to yield for car B and then nudge around the parking car, designated by the arrow. (d \& e): Interaction between two vehicles at an intersection where can A is predicted to yield for car B or make a right-turn behind. After car B passes, car A can make a left turn. Also, note the bimodal prediction of the pedestrian that is located at the corner. (f and g): Predictions in a parking lot and atypical roadgraph.}
    \label{fig:qualexamples_womd}
\end{figure}

\begin{figure}[p!]
    \centering
    \begin{tabular}{ccc}
    \includegraphics[width=0.3\columnwidth]{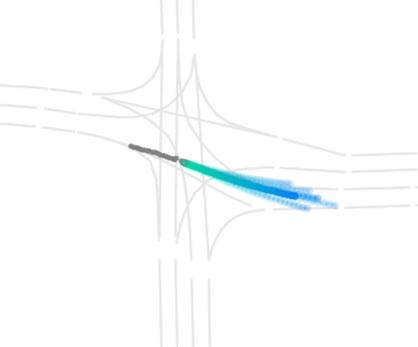} &
    \includegraphics[width=0.3\columnwidth]{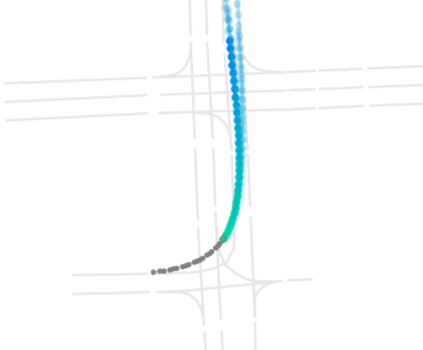} &
    \includegraphics[scale=0.4]{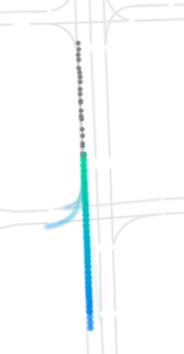} \\
    (a) & (b) & (c) \\ 
    \includegraphics[width=0.3\columnwidth]{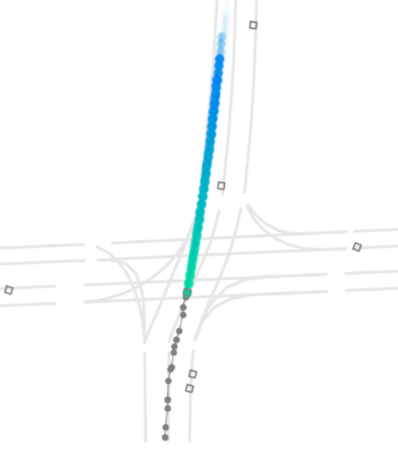} &
    \includegraphics[width=0.3\columnwidth]{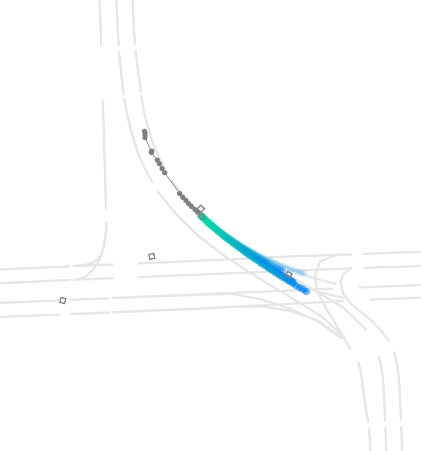} &
    \includegraphics[width=0.3\columnwidth]{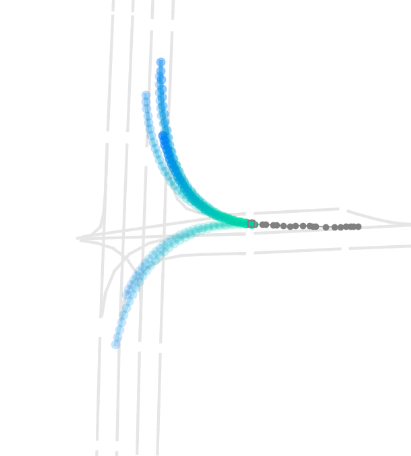} \\
    (d) & (e) & (f) \\
    \end{tabular}
    \caption{Examples of \modelname predictions for 8 seconds in Argoverse scenes, showing the ability to follow different lane geometries.}
    \label{fig:qualexamples_argoverse}
\end{figure}

Figure \ref{fig:qualexamples_womd} shows examples of Multipath++ on WOMD scenes. 
Figure \ref{fig:qualexamples_argoverse} shows examples of Multipath++ on Argoverse scenes. These examples show the ability of \modelname to handle different road layouts and
agent interactions.

\subsection{Ablation Study}
In this section we evaluate our design choices through an ablation study.  Table \ref{tab:ablation} summarizes ablation results. 
In the following subsections we discuss how our architecture choices affect the model performance. 

\subsubsection{Set Functions}
Recall that \modelname uses two types of set functions. Invariant set functions 
are used to encode a set of elements (e.g. agents, roadgraph segments) into a single
feature vector. Equivariant set functions are used to convert the set of learned 
anchors, together with the encoded feature vector as a context, into a corresponding set 
of trajectories with likelihoods. 

We use multi-context gating to represent both types of functions. We experimented with other representations of set functions:
\begin{itemize}
    \item MLP+MaxPool: 
         In this experiment, we replace the multi-context gating (MCG) road network encoder with a MLP+MaxPool applied on points rather than polylines, inspired by PointNet~\cite{qi2017pointnet}. We use a 5 layer deep MLP and RELU activations.
    \item Equivariant DeepSet~\cite{zaheer17deepset}:
        The equivariant set function is represented as a series of
        blocks, each involving an element-wise transformation followed by pooling to compute the context. Unlike MCG, it does not use gating (pointwise multiplication) between set elements and the context vector. Instead, a linear transformation of the context is added to each element. We use a DeepSet of 5 blocks in the predictor.
    \item Transformers~\cite{lee19settransformer}:
        We replace the gating mechanism (element-wise multiplication) on polylines with self-attention. For decoding, we used cross attention where the queries are the learned embeddings and the keys are the various encoder features.
\end{itemize}

\subsubsection{Trajectory representation}
As mentioned in Section \ref{sec:traj_decoding}, we experiment with predicting polynomial 
coefficients for the trajectory, as well predicting kinematic control signals (acceleration and heading change rate). 
We found that polynomial representations hurt performance, counter to conclusions made in PLOP~\cite{buhet20_plop}, where they demonstrated improvements over the then state of the art on PRECOG\cite{precog} and nuScenes\cite{caesar2020nuscenes} using polynomials to represent output trajectories. Furthermore, in the PLOP datasets, we need to predict 4s into the future which is much shorter than our prediction horizon of 10s. For such short futures, polynomial representations are more suitable.  In our case, we do not see much gains from using the polynomial representation, possibly due to the larger dataset size and longer-term prediction horizon.  

The controls-based output works better in distance metrics than a polynomial representations, which suggests it is a more beneficial and domain-specific form of inductive bias. Overall, our results suggest that the simple sequence of raw coordinates trajectory representation works best for distance-based metrics.  However, these unconstrained representations have a non-trivial rate of kinematic infeasibility (TRI-x metrics in Table~\ref{tab:ablation}). Kinematic feasibility and consistency between headings and positions is crucial in practice when such behavior models are used for planning and controls of a real-world robot, an issue that is not captured by public benchmark metrics.

\subsubsection{Ensembling} We explore ensembling,  producing an over-complete set of trajectories that is then summarized using the aggregation proposed in Section~\ref{sec:agg}, as well as their combination. The number of ensembles is denoted by $E$ and the number of trajecctories per ensemble is denoted by $L$. Finally we aggregate the $E \cdot L$ trajectories to $M=6$ which is the required number of trajectories for the WOMD submission.

\subsubsection{Anchor representation} We explore learned and kmeans based anchor representation.

\begin{table}
\centering
{\small
\begin{tabular}{l|ccccccc}
\hline
                      & minDE & minADE  & MR  & AUC & TRI-h (\%) & TRI-c (\%) & TRI-hc (\%)\\
\hline
Original MultiPath & 4.752 & 1.796 & 0.749 & -- & -- & -- & --\\
\hline
\multicolumn{5}{c}{Set Function} \\
\hline
MLP+MaxPool & 2.693 & 1.107 & 0.528  & 0.367 & -- & -- & -- \\
DeepSet & 2.562 & 1.055 & 0.5 & 0.368 & -- & -- & -- \\
Transformer & 2.479 & 1.042 & 0.479 & 0.3687 & -- & -- & -- \\
1 MCG block & 2.764 & 1.15 & 0.55 & 0.312 & -- & -- & -- \\
5 stacked MCG blocks$^\star$ & \bf 2.305 & \bf 0.978 & \bf 0.44 & \bf 0.393 & -- & -- & -- \\
\hline
\multicolumn{5}{c}{Trajectory Representation} \\
\hline
Polynomial & 2.537 & 1.041  & 0.501 & 0.368 & n/a & 1.92 & n/a \\
Control & 2.319 & 0.987  & 0.449 & 0.386 & \bf 0.00 & 1.22 & \bf 0.00 \\
Raw coordinates & \bf 2.305 & \bf 0.978 & \bf 0.44 & \bf 0.393 & n/a & 1.08 & n/a\\
Raw coordinates w/ heading$^\star$ &  2.311 & \bf 0.978 &  0.443 & 0.395 & 4.10 & \bf 1.04 & 9.92\\
\hline
\multicolumn{5}{c}{Ensembling} \\
\hline
$E=1, L=6$ & 2.333 & 0.982 & 0.410 & 0.240 & -- & -- & -- \\
$E=5, L=6$ & \bf 2.18 & \bf 0.948 & \bf 0.395 &0.297 & -- & -- & -- \\
$E=1, L=64$ & 2.487 & 1.057 & 0.473 & 0.367 & -- & -- & -- \\
$E=5, L=64$ $^\star$ & 2.305 & 0.978 & 0.44 & \bf 0.393 & -- & -- & -- \\
\hline
\multicolumn{4}{c}{Anchors} \\
\hline
Static k-means anchors & 2.99 & 1.22 & 0.578 & 0.324 & -- & -- & -- \\
Learned anchors$^\star$ & \bf 2.305 & \bf  0.978 &\bf  0.44 &\bf  0.393 & -- & -- & -- \\
\hline
\end{tabular}
}
\caption{Model ablation on WOMD validatation set. Metrics are parameterized by $k=6$, $t=8s$, and $d=2m$. 
\newline $^\star$ denotes the reference configuration: road encoding, state history encoding and interaction encoding as described in Section~\ref{sec:model}. ``n/a'' denotes a model that does not predict heading.}
\label{tab:ablation}
\end{table}

\begin{figure*}
    \centering
    \includegraphics[width=1.0\textwidth]{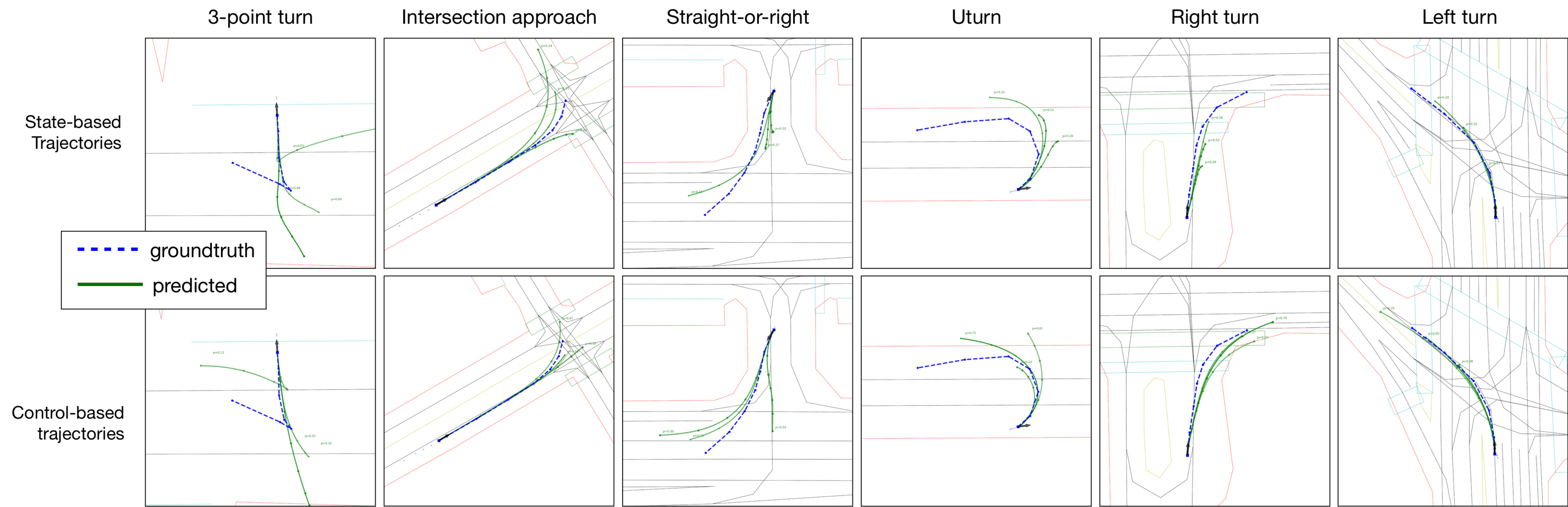}
    \caption{{\small Qualitative results showing the effect of control-based trajectory representation. Top row: \modelname with state-based trajectories.  Bottom row: \modelname with control-based trajectories.}}
    \label{fig:qual}
\end{figure*}

\begin{figure}
\begin{subfigure}{\ }
  \centering
  \includegraphics[width=.8\linewidth]{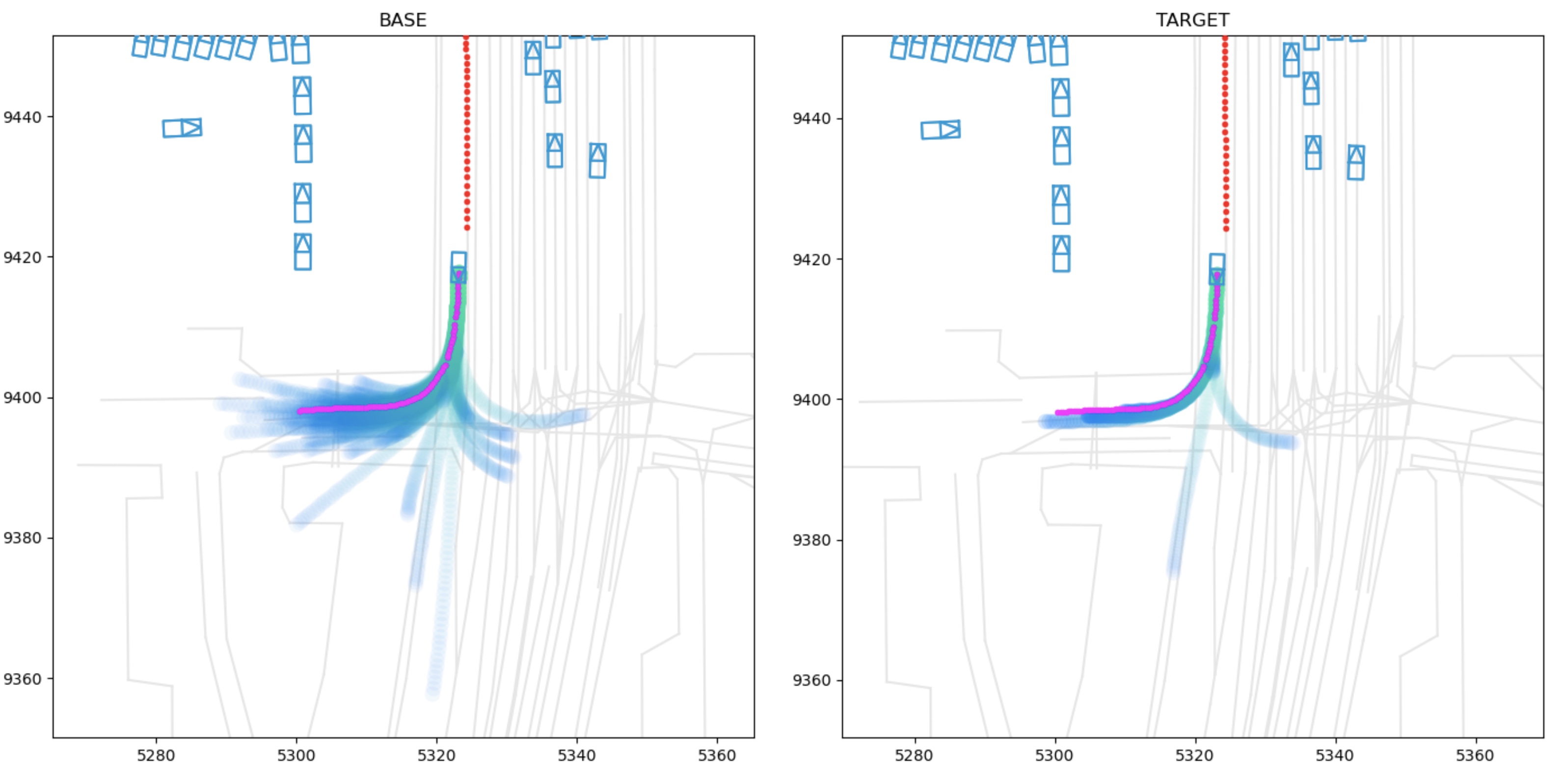}
  \caption{Example of improved diversity due to aggregation. }
\end{subfigure}

\begin{subfigure}{\ }
  \centering
  \includegraphics[width=.8\linewidth]{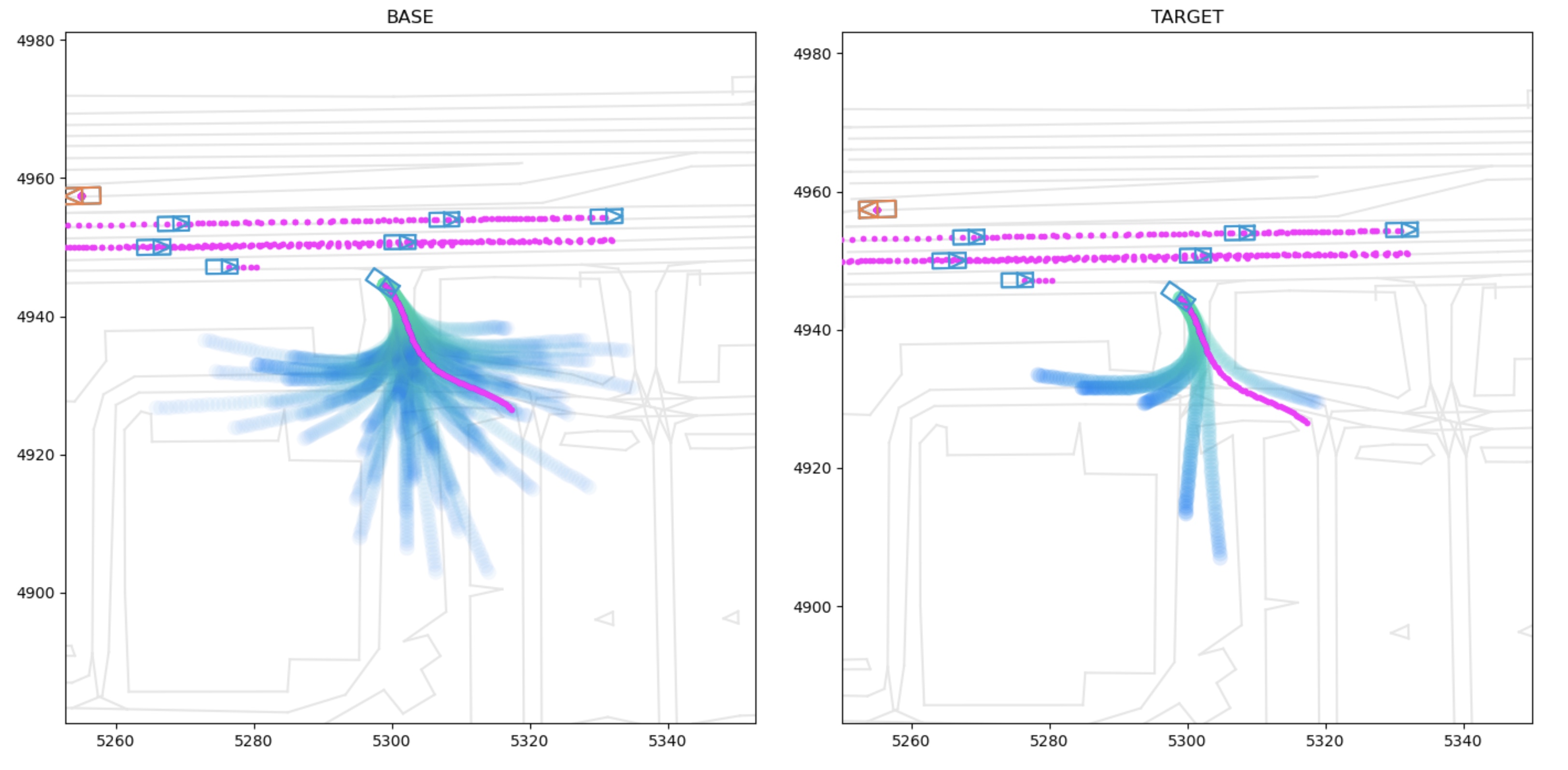}
  \caption{Illustration of weeding out unrealistic trajectories after aggregation.}
\end{subfigure}

\begin{subfigure}{\ }
  \centering
  \includegraphics[width=.8\linewidth]{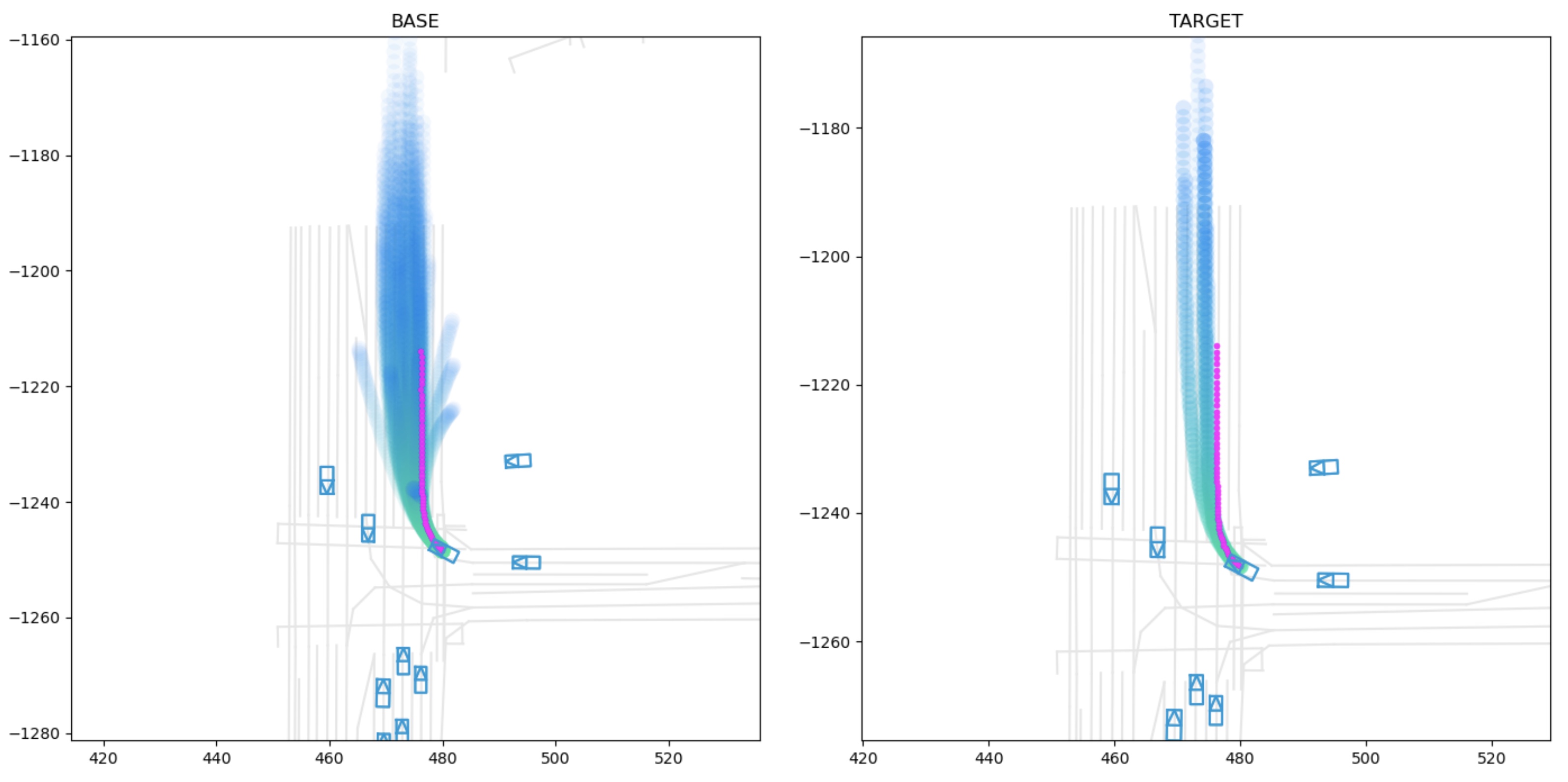}
  \caption{Illustration of improved lane diversity: The model is able to predict the agent going to multiple lanes after aggregation.}
\end{subfigure}

\label{fig:aggregation}
\end{figure}

\subsection{Discussion}

First, we remark that \modelname is a significant improvement over its predecessor MultiPath, as seen in Tables~\ref{tbl:womd} and~\ref{tab:ablation}. As discussed in this paper, they differ in many design dimensions, the primary being the change from a dense top-down raster representation to a sparse, element-based representation with agent-centric coordinate systems.  Other design choices are validated in isolation in the following discussion.

We find that MLP+MaxPool performs the worst among all set function variants as expected due to limited capacity. DeepSet is able to outperform MLP+MaxPool.  Also increasing the depth of the MCG gives consistently better results owing to effective increase in capacity and flow of information across skip connections.  We get the best performance by increasing the depth of the MCG to 5 layers.
 
 We find that learning anchors (``Learned anchors'') is more effective than using a set of anchors obtained a priori via k-means. This runs counter to the original findings in the MultiPath paper~\cite{sapp2019multipath} that anchor-free models suffer from mode collapse. The difference could possibly be due to the richer and more structured inputs, improved model architecture, and larger batch sizes in \modelname. We leave more detailed ablations on this issue between the two approaches to future work. 
t
We compare the baseline of directly outputting a single head with 6 trajectories ($E=1, L=6$), to training 5 ensemble heads ($E=5, L=6$). We see that ensembling significantly improves most metrics, and particularly minDE, for which this combination is best.  We also train a model with a single head that outputs 64 trajectories, followed by our aggregation method that reduces them to 6 ($E=1, L=64$). Compared to our initial baseline, this model significantly improves $MR$ and $AUC$ that require diverse predictions, but regresses the average trajectory distance metrics $minDE$, and even $minADE$ a little bit. This suggests that the different metrics pose different solution requirements. As expected, our aggregation criterion is well suited to preserving diversity, while straight-up ensembling is better at capturing the average distribution. Finally, our experiment ($E=5, L=64$) with more ensemble heads and more predictions per ensemble combines the strengths of both techniques, obtaining a strictly superior performance in all metrics compared to the baseline.

\subsection{Conclusion} 
We proposed a novel behavior prediction system, \modelname, by carefully considering choices for input representation and encoding, fusing encodings, and representing the output distribution. We demonstrated state-of-the-art results on popular benchmarks for behavior prediction. Furthermore, we surveyed existing methods, analyzed our approach empirically, and provided practical insights for the research community. In particular, we showed the importance of sparse encoding, efficient fusion methods, control-based methods, and learned anchors. Finally, we provided a practical guide for various tricks used for training and inference to improve robustness, increase diversity, handle missing data, and ensure fast convergence during training.

{\small
\bibliographystyle{plain}
\bibliography{references}

\begin{thebibliography}{10}

\bibitem{sociallstm}
Alexandre "Alahi, Kratarth Goel, Vignesh Ramanathan, Alexandre Robicquet,
  Li~Fei-Fei, and Silvio Savarese.
\newblock {Social LSTM: Human Trajectory Prediction in Crowded Spaces}.
\newblock In {\em CVPR}, 2016.

\bibitem{biktairov2020prank}
Yuriy Biktairov, Maxim Stebelev, Irina Rudenko, Oleh Shliazhko, and Boris
  Yangel.
\newblock Prank: motion prediction based on ranking.
\newblock {\em arXiv preprint arXiv:2010.12007}, 2020.

\bibitem{bishop2006pattern}
Christopher~M Bishop.
\newblock {\em Pattern recognition and machine learning}.
\newblock springer, 2006.

\bibitem{buhet20_plop}
Thibault Buhet, {\'E}.~Wirbel, and Xavier Perrotton.
\newblock Plop: Probabilistic polynomial objects trajectory planning for
  autonomous driving.
\newblock {\em ArXiv}, abs/2003.08744, 2020.

\bibitem{buhet2020plop}
Thibault Buhet, Emilie Wirbel, and Xavier Perrotton.
\newblock Plop: Probabilistic polynomial objects trajectory planning for
  autonomous driving.
\newblock {\em arXiv preprint arXiv:2003.08744}, 2020.

\bibitem{nuscenes2019}
Holger Caesar, Varun Bankiti, Alex~H. Lang, Sourabh Vora, Venice~Erin Liong,
  Qiang Xu, Anush Krishnan, Yu~Pan, Giancarlo Baldan, and Oscar Beijbom.
\newblock nuscenes: A multimodal dataset for autonomous driving.
\newblock {\em arXiv preprint arXiv:1903.11027}, 2019.

\bibitem{caesar2020nuscenes}
Holger Caesar, Varun Bankiti, Alex~H. Lang, Sourabh Vora, Venice~Erin Liong,
  Qiang Xu, Anush Krishnan, Yu~Pan, Giancarlo Baldan, and Oscar Beijbom.
\newblock nuscenes: A multimodal dataset for autonomous driving, 2020.

\bibitem{carion20detr}
Nicolas Carion, Francisco Massa, Gabriel Synnaeve, Nicolas Usunier, Alexander
  Kirillov, and Sergey Zagoruyko.
\newblock End-to-end object detection with transformers.
\newblock In Andrea Vedaldi, Horst Bischof, Thomas Brox, and Jan-Michael Frahm,
  editors, {\em Computer Vision -- ECCV 2020}, pages 213--229, Cham, 2020.
  Springer International Publishing.

\bibitem{casas2020spagnn}
Sergio Casas, Cole Gulino, Renjie Liao, and Raquel Urtasun.
\newblock Spagnn: Spatially-aware graph neural networks for relational behavior
  forecasting from sensor data.
\newblock In {\em IEEE Intl. Conf. on Robotics and Automation}. IEEE, 2020.

\bibitem{casas2018intentnet}
Sergio Casas, Wenjie Luo, and Raquel Urtasun.
\newblock Intentnet: Learning to predict intention from raw sensor data.
\newblock In {\em Conf. on Robot Learning}, 2018.

\bibitem{casas2021mp3}
Sergio Casas, Abbas Sadat, and Raquel Urtasun.
\newblock Mp3: A unified model to map, perceive, predict and plan.
\newblock In {\em Proceedings of the IEEE/CVF Conference on Computer Vision and
  Pattern Recognition}, pages 14403--14412, 2021.

\bibitem{chang2019argoverse}
Ming-Fang Chang, John Lambert, Patsorn Sangkloy, Jagjeet Singh, Slawomir Bak,
  Andrew Hartnett, De~Wang, Peter Carr, Simon Lucey, Deva Ramanan, et~al.
\newblock Argoverse: 3d tracking and forecasting with rich maps.
\newblock In {\em CVPR}, 2019.

\bibitem{cui2020deep}
Henggang Cui, Thi Nguyen, Fang-Chieh Chou, Tsung-Han Lin, Jeff Schneider, David
  Bradley, and Nemanja Djuric.
\newblock Deep kinematic models for kinematically feasible vehicle trajectory
  predictions.
\newblock In {\em IEEE Intl. Conf. on Robotics and Automation}, pages
  10563--10569. IEEE, 2020.

\bibitem{cui2019multimodal}
Henggang Cui, Vladan Radosavljevic, Fang-Chieh Chou, Tsung-Han Lin, Thi Nguyen,
  Tzu-Kuo Huang, Jeff Schneider, and Nemanja Djuric.
\newblock Multimodal trajectory predictions for autonomous driving using deep
  convolutional networks.
\newblock In {\em IEEE Intl. Conf. on Robotics and Automation}, 2019.

\bibitem{dai2021coatnet}
Zihang Dai, Hanxiao Liu, Quoc~V Le, and Mingxing Tan.
\newblock Coatnet: Marrying convolution and attention for all data sizes.
\newblock {\em arXiv preprint arXiv:2106.04803}, 2021.

\bibitem{dempster77}
A.~P. Dempster, N.~M. Laird, and D.~B. Rubin.
\newblock Maximum likelihood from incomplete data via the em algorithm.
\newblock {\em Journal of the Royal Statistical Society. Series B
  (Methodological)}, 39(1):1--38, 1977.

\bibitem{dosovitskiy2020ViT}
Alexey Dosovitskiy, Lucas Beyer, Alexander Kolesnikov, Dirk Weissenborn,
  Xiaohua Zhai, Thomas Unterthiner, Mostafa Dehghani, Matthias Minderer, Georg
  Heigold, Sylvain Gelly, et~al.
\newblock An image is worth 16x16 words: Transformers for image recognition at
  scale.
\newblock In {\em International Conference on Learning Representations}, 2020.

\bibitem{ettinger2021womd}
Scott Ettinger, Shuyang Cheng, Benjamin Caine, Chenxi Liu, Hang Zhao, Sabeek
  Pradhan, Yuning Chai, Ben Sapp, Charles Qi, Yin Zhou, Zoey Yang, Aurelien
  Chouard, Pei Sun, Jiquan Ngiam, Vijay Vasudevan, Alexander McCauley, Jonathon
  Shlens, and Dragomir Anguelov.
\newblock Large scale interactive motion forecasting for autonomous driving :
  The waymo open motion dataset, 2021.

\bibitem{eslbook}
Jerome~H Friedman.
\newblock {\em The elements of statistical learning: Data mining, inference,
  and prediction}.
\newblock springer open, 2017.

\bibitem{gao2020vectornet}
Jiyang Gao, Chen Sun, Hang Zhao, Yi~Shen, Dragomir Anguelov, Congcong Li, and
  Cordelia Schmid.
\newblock {VectorNet}: Encoding hd maps and agent dynamics from vectorized
  representation.
\newblock In {\em CVPR}, 2020.

\bibitem{gilles2021gohome}
Thomas Gilles, Stefano Sabatini, Dzmitry Tsishkou, Bogdan Stanciulescu, and
  Fabien Moutarde.
\newblock Gohome: Graph-oriented heatmap output for future motion estimation.
\newblock {\em arXiv preprint arXiv:2109.01827}, 2021.

\bibitem{gilles21home}
Thomas Gilles, Stefano Sabatini, Dzmitry Tsishkou, Bogdan Stanciulescu, and
  Fabien Moutarde.
\newblock {HOME:} heatmap output for future motion estimation.
\newblock {\em CoRR}, abs/2105.10968, 2021.

\bibitem{gu21dense_tnt}
Junru Gu, Qiao Sun, and Hang Zhao.
\newblock Densetnt: Waymo open dataset motion prediction challenge 1st place
  solution.
\newblock {\em CoRR}, abs/2106.14160, 2021.

\bibitem{SocialGAN}
Agrim Gupta, Justin Johnson, Li~Fei-Fei, Silvio Savarese, and Alexandre Alahi.
\newblock Social {GAN}: Socially acceptable trajectories with generative
  adversarial networks.
\newblock In {\em CVPR}, 2018.

\bibitem{ResNet16}
Kaiming He, Xiangyu Zhang, Shaoqing Ren, and Jian Sun.
\newblock Deep residual learning for image recognition.
\newblock In {\em CVPR}, 2016.

\bibitem{homayounfar2018maps}
Namdar Homayounfar, Wei-Chiu Ma, Shrinidhi~Kowshika Lakshmikanth, and Raquel
  Urtasun.
\newblock Hierarchical recurrent attention networks for structured online maps.
\newblock In {\em Proceedings of the IEEE Conference on Computer Vision and
  Pattern Recognition}, pages 3417--3426, 2018.

\bibitem{hong2019rules}
Joey Hong, Benjamin Sapp, and James Philbin.
\newblock Rules of the road: Predicting driving behavior with a convolutional
  model of semantic interactions.
\newblock In {\em CVPR}, 2019.

\bibitem{wimp2020}
Siddhesh Khandelwal, William Qi, Jagjeet Singh, Andrew Hartnett, and Deva
  Ramanan.
\newblock What-if motion prediction for autonomous driving.
\newblock {\em ArXiv}, 2020.

\bibitem{lee19settransformer}
Juho Lee, Yoonho Lee, Jungtaek Kim, Adam~R. Kosiorek, Seungjin Choi, and
  Yee~Whye Teh.
\newblock Set transformer: {A} framework for attention-based
  permutation-invariant neural networks.
\newblock In Kamalika Chaudhuri and Ruslan Salakhutdinov, editors, {\em
  Proceedings of the 36th International Conference on Machine Learning, {ICML}
  2019, 9-15 June 2019, Long Beach, California, {USA}}, volume~97 of {\em
  Proceedings of Machine Learning Research}, pages 3744--3753. {PMLR}, 2019.

\bibitem{DESIRE}
Namhoon Lee, Wongun Choi, Paul Vernaza, Christopher~B Choy, Philip H~S Torr,
  and Manmohan Chandraker.
\newblock {DESIRE}: Distant future prediction in dynamic scenes with
  interacting agents.
\newblock In {\em CVPR}, 2017.

\bibitem{liang2020garden}
Junwei Liang, Lu~Jiang, Kevin Murphy, Ting Yu, and Alexander Hauptmann.
\newblock The garden of forking paths: Towards multi-future trajectory
  prediction.
\newblock In {\em CVPR}, 2020.

\bibitem{liang2020laneGCN}
Ming Liang, Bin Yang, Rui Hu, Yun Chen, Renjie Liao, Song Feng, and Raquel
  Urtasun.
\newblock Learning lane graph representations for motion forecasting.
\newblock {\em arXiv preprint arXiv:2007.13732}, 2020.

\bibitem{marchetti2020mantra}
Francesco Marchetti, Federico Becattini, Lorenzo Seidenari, and Alberto~Del
  Bimbo.
\newblock Mantra: Memory augmented networks for multiple trajectory prediction.
\newblock In {\em Proceedings of the IEEE/CVF Conference on Computer Vision and
  Pattern Recognition}, pages 7143--7152, 2020.

\bibitem{mercat2020multi}
Jean Mercat, Thomas Gilles, Nicole Zoghby, Guillaume Sandou, Dominique
  Beauvois, and Guillermo Gil.
\newblock Multi-head attention for joint multi-modal vehicle motion
  forecasting.
\newblock In {\em IEEE Intl. Conf. on Robotics and Automation}, 2020.

\bibitem{ngiam21scene_transformer}
Jiquan Ngiam, Benjamin Caine, Vijay Vasudevan, Zhengdong Zhang,
  Hao{-}Tien~Lewis Chiang, Jeffrey Ling, Rebecca Roelofs, Alex Bewley, Chenxi
  Liu, Ashish Venugopal, David Weiss, Benjamin Sapp, Zhifeng Chen, and Jonathon
  Shlens.
\newblock Scene transformer: {A} unified multi-task model for behavior
  prediction and planning.
\newblock {\em CoRR}, abs/2106.08417, 2021.

\bibitem{park2020diverse}
Seong~Hyeon Park, Gyubok Lee, Manoj Bhat, Jimin Seo, Minseok Kang, Jonathan
  Francis, Ashwin~R Jadhav, Paul~Pu Liang, and Louis-Philippe Morency.
\newblock Diverse and admissible trajectory forecasting through multimodal
  context understanding.
\newblock {\em arXiv preprint arXiv:2003.03212}, 2020.

\bibitem{phan2019covernet}
Tung Phan-Minh, Elena~Corina Grigore, Freddy~A Boulton, Oscar Beijbom, and
  Eric~M Wolff.
\newblock {CoverNet}: Multimodal behavior prediction using trajectory sets.
\newblock {\em arXiv:1911.10298}, 2019.

\bibitem{qi2017pointnet}
Charles~R Qi, Hao Su, Kaichun Mo, and Leonidas~J Guibas.
\newblock Pointnet: Deep learning on point sets for 3d classification and
  segmentation.
\newblock In {\em CVPR}, 2017.

\bibitem{Refaat2019AgentPF}
Khaled~S. Refaat, Kai Ding, Natalia Ponomareva, and St{\'e}phane Ross.
\newblock Agent prioritization for autonomous navigation.
\newblock {\em 2019 IEEE/RSJ International Conference on Intelligent Robots and
  Systems (IROS)}, pages 2060--2067, 2019.

\bibitem{rhinehart2018r2p2}
Nicholas Rhinehart, Kris Kitani, and Paul Vernaza.
\newblock {R2P2}: A reparameterized pushforward policy for diverse, precise
  generative path forecasting.
\newblock In {\em ECCV}, 2018.

\bibitem{precog_Rhinehart_2019_ICCV}
Nicholas Rhinehart, Rowan McAllister, Kris Kitani, and Sergey Levine.
\newblock {PRECOG}: Prediction conditioned on goals in visual multi-agent
  settings.
\newblock In {\em Intl. Conf. on Computer Vision}, 2019.

\bibitem{rhinehart2019precog}
Nicholas Rhinehart, Rowan McAllister, Kris Kitani, and Sergey Levine.
\newblock Precog: Prediction conditioned on goals in visual multi-agent
  settings.
\newblock In {\em ECCV}, 2019.

\bibitem{precog}
Nicholas Rhinehart, Rowan McAllister, Kris~M. Kitani, and Sergey Levine.
\newblock {PRECOG:} prediction conditioned on goals in visual multi-agent
  settings.
\newblock {\em CoRR}, abs/1905.01296, 2019.

\bibitem{salzmann2020trajectron++}
Tim Salzmann, Boris Ivanovic, Punarjay Chakravarty, and Marco Pavone.
\newblock Trajectron++: Multi-agent generative trajectory forecasting with
  heterogeneous data for control.
\newblock {\em arXiv preprint arXiv:2001.03093}, 2020.

\bibitem{sapp2019multipath}
Benjamin Sapp, Yuning Chai, Mayank Bansal, and Dragomir Anguelov.
\newblock {MultiPATH}: Multiple probabilistic anchor trajectory hypotheses for
  behavior prediction.
\newblock In {\em Conf. on Robot Learning}, 2019.

\bibitem{tang_multifuture}
Yichuan~Charlie Tang and Ruslan Salakhutdinov.
\newblock Multiple futures prediction.
\newblock In {\em Proceedings of the 33rd International Conference on Neural
  Information Processing Systems}, Red Hook, NY, USA, 2019. Curran Associates
  Inc.

\bibitem{tolstaya2021cbp}
Ekaterina~I. Tolstaya, Reza Mahjourian, Carlton Downey, Balakrishnan
  Varadarajan, Benjamin Sapp, and Dragomir Anguelov.
\newblock Identifying driver interactions via conditional behavior prediction.
\newblock In {\em {IEEE} International Conference on Robotics and Automation,
  {ICRA} 2021, Xi'an, China, May 30 - June 5, 2021}, pages 3473--3479. {IEEE},
  2021.

\bibitem{vaswani2017attention}
Ashish Vaswani, Noam Shazeer, Niki Parmar, Jakob Uszkoreit, Llion Jones,
  Aidan~N Gomez, {\L}ukasz Kaiser, and Illia Polosukhin.
\newblock Attention is all you need.
\newblock In {\em NeurIPS}, 2017.

\bibitem{ye2021tpcn}
Maosheng Ye, Tongyi Cao, and Qifeng Chen.
\newblock Tpcn: Temporal point cloud networks for motion forecasting.
\newblock In {\em Proceedings of the IEEE/CVF Conference on Computer Vision and
  Pattern Recognition}, pages 11318--11327, 2021.

\bibitem{kitani_diverse_forecasting_dpps}
Ye~Yuan and Kris Kitani.
\newblock Diverse trajectory forecasting with determinantal point processes.
\newblock {\em ICLR}, 2020.

\bibitem{zaheer17deepset}
Manzil Zaheer, Satwik Kottur, Siamak Ravanbakhsh, Barnabas Poczos, Russ~R
  Salakhutdinov, and Alexander~J Smola.
\newblock Deep sets.
\newblock In I.~Guyon, U.~V. Luxburg, S.~Bengio, H.~Wallach, R.~Fergus,
  S.~Vishwanathan, and R.~Garnett, editors, {\em Advances in Neural Information
  Processing Systems}, volume~30. Curran Associates, Inc., 2017.

\bibitem{neural_motion_planner_zeng2019}
Wenyuan Zeng, Wenjie Luo, Simon Suo, Abbas Sadat, Bin Yang, Sergio Casas, and
  Raquel Urtasun.
\newblock End-to-end interpretable neural motion planner.
\newblock In {\em CVPR}, 2019.

\bibitem{interactiondataset}
Wei Zhan, Liting Sun, Di~Wang, Haojie Shi, Aubrey Clausse, Maximilian Naumann,
  Julius K\"ummerle, Hendrik K\"onigshof, Christoph Stiller, Arnaud
  de~La~Fortelle, and Masayoshi Tomizuka.
\newblock {INTERACTION} {Dataset}: {An} {INTERnational}, {Adversarial} and
  {Cooperative} {moTION} {Dataset} in {Interactive} {Driving} {Scenarios} with
  {Semantic} {Maps}.
\newblock {\em arXiv:1910.03088}, 2019.

\bibitem{zhao2020tnt}
Hang Zhao, Jiyang Gao, Tian Lan, Chen Sun, Benjamin Sapp, Balakrishnan
  Varadarajan, Yue Shen, Yi~Shen, Yuning Chai, Cordelia Schmid, et~al.
\newblock Tnt: Target-driven trajectory prediction.
\newblock {\em arXiv preprint arXiv:2008.08294}, 2020.

\bibitem{zwicker2001surface_splat}
Matthias Zwicker, Hanspeter Pfister, Jeroen Van~Baar, and Markus Gross.
\newblock Surface splatting.
\newblock In {\em Proceedings of the 28th annual conference on Computer
  graphics and interactive techniques}, pages 371--378, 2001.

\end{thebibliography}
}
\appendix
\section{Details and Derivation of Aggregation Algorithm}
\label{sec:appendix-aggregation}

By having an overcomplete trajectory representation that is later aggregated into a fixed small number of trajectories, we attempt to address two kinds of uncertainties in the data:
\begin{itemize}
    \item \emph{Aleatoric uncertainty}: This is a natural variation in the data. For example an agent can either take a left or right turn or change lanes, etc given the same context information. This level of ambiguity cannot be resolved by increasing the model capacity, but rather the model needs to predict calibrated probabilities for these outcomes. Despite the theoretical possibility of modeling these variations using a small number of output trajectories directly, there are  several challenges in learning. Some examples include mode collapse and failure to model these variations due to limited model capacity. Training the model to produce an overcomplete representation forces the model to output a diverse distribution of trajectories and could make it more resistant to mode collapse. Following this up with greedy iterative trajectory aggregation enhances diversity in the final output.
    \item \emph{Epistemic uncertainty}: This is the variation across model outputs, which typically indicates the model's failure to capture certain aspects of the scene or input features. Such variations could occur if some models are poorly trained or haven't seen a particular slice of the data. By doing model ensembling, we attempt to reduce this uncertainty.
\end{itemize}

For ease of exposition, we assume each to trajectory to be composed of a single time point; the same computations are applied to each time step in a future sequence.
The output $\Psi$ is a Gaussian mixture model (GMM) distribution with $M'$ modes on the future position:
\begin{equation}
\mathbf{p}(x,y; \Psi)  = \sum_{j=1}^{M'} q_j \mathcal{N}(x, y; \mathbf{\mu}_j, \mathbf{\Sigma}_j)
\end{equation}

We  formulate  the  aggregation as obtaining an $M$-mode GMM $\bar{\mathbf{\Psi}}$ which minimizes the KL-divergence $D_\mathrm{KL}(\mathbf{\Psi} || \bar{\mathbf{\Psi}})$.
This is equivalent to maximizing the expected log likelihood $\mathbf{p}(x,y; \bar{\operatorname{\Psi}})$ of a sample point $(x,y)$ drawn from the overcomplete distribution $\Psi$:
\begin{equation}
\label{eq:agg_ll}
f(\bar{\Psi}) = \int_{\mathbf{x}} p(\mathbf{x}; \Psi) \log p(\mathbf{x};  \bar{\Psi}) d\mathbf{x}
\end{equation}

Assuming the overcomplete distribution approximates the real distribution, this is roughly equivalent to fitting the compact distribution to the real data,  but with the added benefits described above.
Directly maximizing \eqref{eq:agg_ll} is intractable. Hence we attempt to employ an Expectation-Maximization-like algorithm to obtain a local maximum. The difference in the objective function between an old and new value may be written as

\begin{equation}
f(\bar{\Psi'}) - f(\bar{\Psi}) =   \int_{\mathbf{x}} p(\mathbf{x}; \Psi)  \log \left[ \frac{p(\mathbf{x}; \bar{\Psi}')}{p(  \mathbf{x}; \bar{\Psi})} \right] d\mathbf{x}
\end{equation}

Denoting the hidden variable h to be a mixture in the compact representation, we may write:
\begin{eqnarray}
 ~ & \log \left[ \frac{p(\mathbf{x}; \bar{\Psi}')}{p(  \mathbf{x}; \bar{\Psi})} \right]\\
  = & \sum_{h} p(h| \mathbf{x}; \bar{\Psi}) \log \left[ \frac{p(\mathbf{x}; \bar{\Psi}')}{p(  \mathbf{x}; \bar{\Psi})} \right]\\
 = & \sum_{h} p(h| \mathbf{x}; \bar{\Psi}) \log  \left[ \frac{p(h, \mathbf{x}; \bar{\Psi}')p(h |   \mathbf{x}; \bar{\Psi})}{p(h| \mathbf{x}; \bar{\Psi}') p(h,   \mathbf{x}; \bar{\Psi})} \right]\\
  = & \sum_{h} p(h| \mathbf{x}; \bar{\Psi}) \log  \left[ \frac{p(h, \mathbf{x}; \bar{\Psi}')}{p(h,   \mathbf{x}; \bar{\Psi})} \right] + \sum_{h} p(h| \mathbf{x}; \bar{\Psi}) \log  \left[ \frac{p(h| \mathbf{x}; \bar{\Psi})}{p(h|   \mathbf{x}; \bar{\Psi}')} \right] \\
 = & \sum_{h} p(h| \mathbf{x}; \bar{\Psi}) \log  \left[ \frac{p(h, \mathbf{x}; \bar{\Psi}')}{p(h,   \mathbf{x}; \bar{\Psi})} \right] + D_\mathrm{KL}(p(h| \mathbf{x}; \bar{\Psi}) \Vert p(h| \mathbf{x}; \bar{\Psi}')  )\\
 \geq &  \sum_{h} p(h| \mathbf{x}; \bar{\Psi}) \log  \left[ \frac{p(h, \mathbf{x}; \bar{\Psi}')}{p(h,   \mathbf{x}; \bar{\Psi})} \right]
\end{eqnarray}

Thus
\begin{equation}
f(\bar{\Psi'}) - f(\bar{\Psi}) \geq  \int_{\mathbf{x}} p(\mathbf{x}; \Psi) \sum_{h} p(h | \mathbf{x}; \bar{\Psi}) \log \left[ \frac{p(h,  \mathbf{x}; \bar{\Psi}')}{p(h,  \mathbf{x}; \bar{\Psi})} \right] d\mathbf{x}
\end{equation}

The right hand side is called the Q function in the EM algorithm. Maximizing the Q function with respect to $\bar{\Psi}'$ ensures that the likelihood increases at least as much when we update the parameters to $\bar{\Psi}'$.  Noting that $p(h,  \mathbf{x}; \bar{\Psi}') = p(h; \bar{\Psi}') p(\mathbf{x};\bar{\Psi}' | h) = \bar{q}_h' \mathcal{N}(\mathbf{x} - \bar{\mathbf{\mu}}_h', \bar{\mathbf{\Sigma}}_h')$ and factoring out the terms independent of $\bar{\Psi}'$, we find the update that maximizes the lower bound to be

\begin{equation}
\bar{\Psi} \leftarrow \underset{\bar{\Psi}'}{\operatorname{argmax}} \int_{\mathbf{x}} p(\mathbf{x}; \Psi) \sum_{h} p(h | \mathbf{x}; \bar{\Psi}) \log \left[ \bar{q}_h' \mathcal{N}(\mathbf{x}- \bar{\mathbf{\mu}}_h', \bar{\mathbf{\Sigma}}_h') \right] d\mathbf{x}
\end{equation}

\begin{equation}
\bar{\Psi} \leftarrow \underset{\bar{\Psi}'}{\operatorname{argmax}} \sum_{i=1}^{M'} q_i \int_{\mathbf{x}} p(\mathbf{x}; \mathbf{\mu}_i, \mathbf{\Sigma}_i) \sum_{h} p(h | \mathbf{x}; \bar{\Psi}) \log \left[ \bar{q}_h' \mathcal{N}(\mathbf{x} - \bar{\mathbf{\mu}}_h', \bar{\mathbf{\Sigma}}_h') \right]d\mathbf{x}
\end{equation}
The second equation follows from the fact that the overcomplete distribution $\Psi$ is a mixture of $M'$ Gaussians. The updates can be solved as follows.

\begin{eqnarray*}
    \bar{q}_h' &\leftarrow& \sum_{i=1}^{M'} q_i \int_{\mathbf{x}} p(\mathbf{x}; \mathbf{\mu}_i, \mathbf{\Sigma}_i)  p(h | \mathbf{x}; \bar{\Psi}) d\mathbf{x}  \\   
    \bar{\mathbf{\mu}}_h' &\leftarrow& \frac{1}{\bar{q}_h} \sum_{i=1}^{M'} q_i \int_{\mathbf{x}} \mathbf{x} p(\mathbf{x}; \mathbf{\mu}_i, \mathbf{\Sigma}_i)  p(h | \mathbf{x}; \bar{\Psi}) d\mathbf{x}  \\   
        \bar{\mathbf{\Sigma}}_h' &\leftarrow& \frac{1}{\bar{q}_h} \sum_{i=1}^{M'} q_i \int_{\mathbf{x}} \mathbf{x} p(\mathbf{x}; \mathbf{\mu}_i, \mathbf{\Sigma}_i)  (\mathbf{x} - \bar{\mathbf{\mu}}_h') (\mathbf{x} - \bar{\mathbf{\mu}}_h')^T p(h | \mathbf{x}; \bar{\Psi}) d\mathbf{x} \\   
\end{eqnarray*}
where $p(h | \mathbf{x}; \bar{\Psi})$ is the posterior probability 
that a given sample $\mathbf{x}$ is sampled from the $h^{th}$ component
of the mixture model specified by $\bar{\Psi}$
(here we use the previous estimate for $\bar{\Psi}$). This can be computed as:
\begin{equation}
    p(h | \mathbf{x}; \bar{\Psi}) = \frac{\bar{q}_h\mathcal{N}(\mathbf{x}; \mathbf{\bar{\mu}}_h, \mathbf{\bar{\Sigma}}_h) }{\sum_{k=1}^M \bar{q}_k\mathcal{N}(\mathbf{x}; \mathbf{\bar{\mu}}_k, \mathbf{\bar{\Sigma}}_k)}
\end{equation}

Notice the resemblence with standard GMM, except where $p(\mathbf{x}; \mathbf{\mu}_i, \mathbf{\Sigma}_i)$ is a dirac delta function  in the standard setting (since the input data in standard GMM is a set of points instead of a distribution).  Unlike standard GMM, these expectations (integrations) in the above EM updates are hard to compute in closed form. Instead we employ the approximation for any function $g(\mathbf{x})$
\begin{eqnarray*}
\mathbb{E}_{\mathbf{x} \sim \mathcal{N}(x; \mathbf{\mu}, \mathbf{\Sigma})} \left[g(\mathbf{x}) p(h | \mathbf{x}; \bar{\Psi})  \right] \\
\quad\quad \approx \mathbb{E}_{\mathbf{x} \sim \mathcal{N}(x; \mathbf{\mu}, \mathbf{\Sigma})} \left[g(\mathbf{x}) p(h | \mathbf{\mu}; \bar{\Psi})  \right] \\
\quad\quad = p(h | \mathbf{\mu}; \bar{\Psi}) \mathbb{E}_{\mathbf{x} \sim \mathcal{N}(x; \mathbf{\mu}, \mathbf{\Sigma})} \left[g(\mathbf{x}) \right].
\end{eqnarray*}
In other words, we assume that the posterior probability of any output cluster only depends on the mean of the overcomplete cluster centroid inside the expectation. This approximation is reasonable since most samples  drawn from  the distribution would be concentrated around  the mean. Furthermore as we increase the number of cluster  centroids in the overcomplete representation, the variance within each overcomplete cluster centroid becomes smaller yielding more focus around  the mean.  The set of updates can now be solved  in closed  form  as follows:

\begin{eqnarray*}
    \bar{q}_h &\leftarrow& \sum_{i=1}^{M'} q_i p(h | \mathbf{\mu}_i; \bar{\Psi}) \\   
    \bar{\mathbf{\mu}}_h &\leftarrow& \frac{1}{\bar{q}_h} \sum_{i=1}^{M'} q_i p(h | \mathbf{\mu}_i; \bar{\Psi})  \mathbf{x} \\   
        \bar{\mathbf{\Sigma}}_h &\leftarrow& \frac{1}{\bar{q}_h} \sum_{i=1}^{M'} q_i  p(h | \mathbf{\mu}_i; \bar{\Psi}) \left[ \mathbf{\Sigma}_i +  (\mathbf{\mu}_i - \bar{\mathbf{\mu}}_h) (\mathbf{\mu}_i - \bar{\mathbf{\mu}}_h)^T \right] \\   
\end{eqnarray*}

Since EM is a local optimization method, careful initialization of GMM parameters is important. 
Our initialization criterion of GMM centroids is to maximize the 
probability that future point lies within $\tau$ distance from 
at least one centroid:
\begin{equation}
  \mathbf{\bar{\mu}}_{1:M} = \underset{\mathbf{\bar{\mu}}_{1:M}}{\operatorname{argmax}} \sum_{i=1}^{M'} q_i\,\, \underset{{\bar{\mu} \in   \mathbf{\bar{\mu}}_{1:M}}}{\operatorname{max}}\,\, \mathbb{I} \left(\Vert \mathbf{{\mu}}_i -\bar{\mu} \Vert_2 \leq \tau \right )
\label{eq:em_init}
\end{equation}
Unfortunately, directly optimizing \eqref{eq:em_init} is NP-hard.
So instead, we select an $M$-sized subset of $\mathbf{{\mu}}_{1:M'}$ in a greedy fashion to maximize \eqref{eq:em_init}\footnote{
Note that this subset selection problem is \emph{submodular}, 
which means that a greedy method is guranteed to achieve at 
least $1 - 1/e$ of the optimal subset value.
}.

\newpage
\section{Multipath baseline system diagram}

\begin{figure*}[h!]
    \centering
    \includegraphics[width=0.95\textwidth]{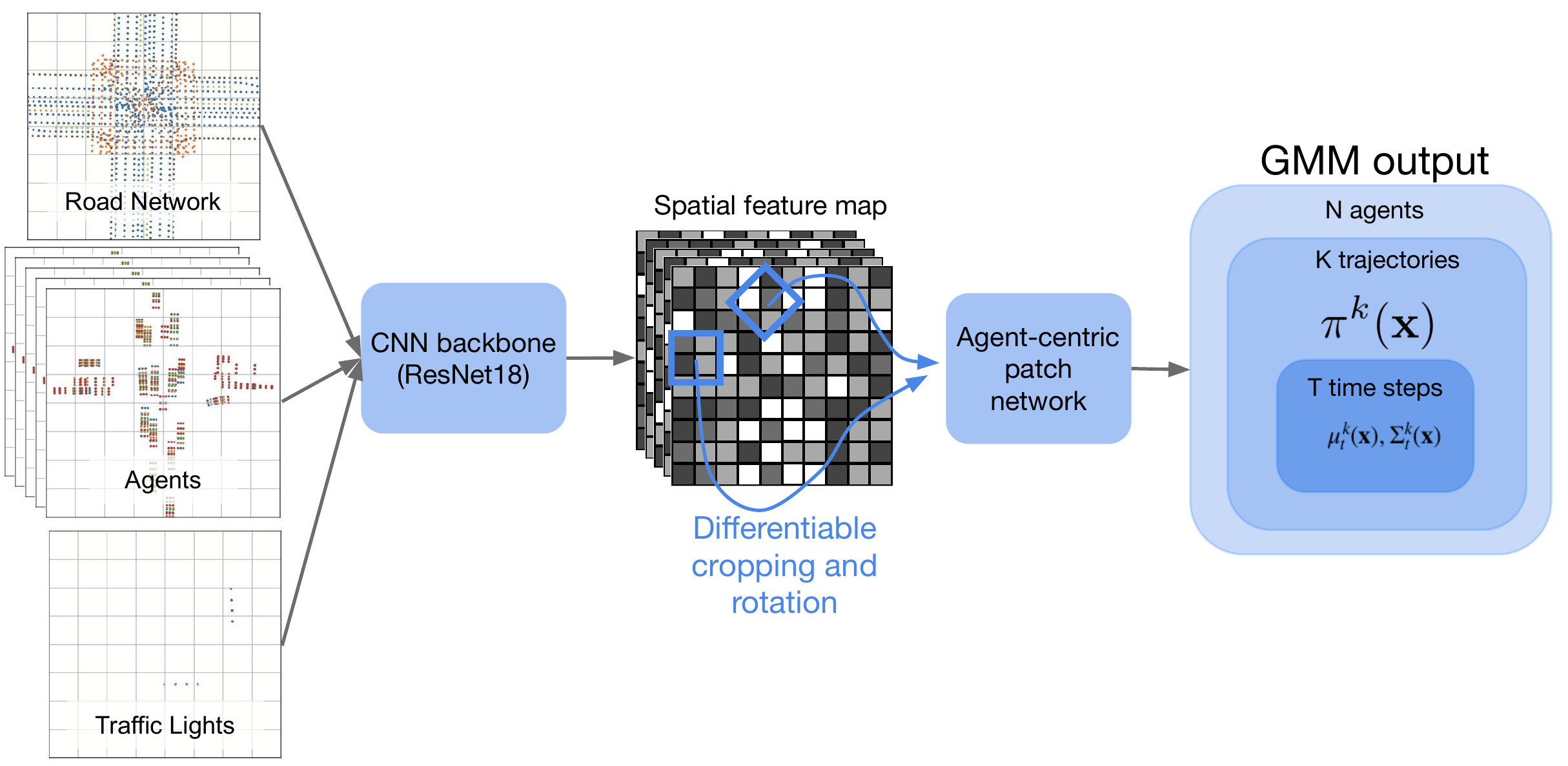}
    \caption{Original MultiPath architecture, with details described in Section~\ref{sec:og_mp}. }
    \label{fig:og_mp}
\end{figure*}

Figure~\ref{fig:og_mp} visualizes the splat rendering rasterization of input points, and the backbone and output architecture.  See Section~\ref{sec:og_mp} for details.

\end{document}